\documentclass{ieeeaccess}
\usepackage{cite}
\usepackage{amsmath,amssymb,amsfonts}
\usepackage{algorithm2e}
\usepackage{graphicx}
\usepackage{textcomp}
\usepackage{multirow}

\definecolor{amber(sae/ece)}{rgb}{1.0, 0.49, 0.0}
\definecolor{amberlite(sae/ece)}{rgb}{1.0, 0.69, 0.2}

\begin{document}

\history{Received January 12, 2021, accepted January 29, 2021}
\doi{10.1109/ACCESS.2021.3056958}

\title{Cylindrical Shape Decomposition for 3D Segmentation of Tubular Objects}

\author{\uppercase{Ali Abdollahzadeh}\authorrefmark{1}, \uppercase{Alejandra Sierra\authorrefmark{1}, and Jussi Tohka}\authorrefmark{1}}

\address[1]{A.I.Virtanen Institute for Molecular Sciences, University of Eastern Finland, Kuopio, Finland}

\tfootnote{This work was supported in part by the Academy of Finland under Grant 316258 and Grant 323385 and in part by the Jane and Aatos Erkko's Foundation.}

\markboth
{A. Abdollahzadeh \headeretal: Cylindrical Shape Decomposition for 3D Segmentation of Tubular Objects}
{A. Abdollahzadeh \headeretal: Cylindrical Shape Decomposition for 3D Segmentation of Tubular Objects}

\corresp{Corresponding author: Ali Abdollahzadeh (e-mail: ali.abdollahzadeh@uef.fi).}

\begin{abstract}
We develop a cylindrical shape decomposition (CSD) algorithm to decompose an object, a union of several tubular structures, into its semantic components. We decompose the object using its curve skeleton and restricted translational sweeps. For that, CSD partitions the curve skeleton into maximal-length sub-skeletons over an orientation cost, each sub-skeleton corresponds to a semantic component. To find the intersection of the tubular components, CSD translationally sweeps the object in decomposition intervals to identify critical points at which the shape of the object changes substantially. CSD cuts the object at critical points and assigns the same label to parts along the same sub-skeleton, thereby constructing a semantic component. The proposed method further reconstructs the acquired semantic components at the intersection of object parts using generalized cylinders. We apply CSD for segmenting axons in large 3D electron microscopy images and decomposing vascular networks and synthetic objects. We show that our proposal is robust to severe surface noise and outperforms state-of-the-art decomposition techniques in its applications.
\end{abstract}


\titlepgskip=-15pt
\maketitle

\section{Introduction}
\PARstart{S}{hape} decomposition is a fundamental problem in geometry processing where an arbitrary object is regarded as an arrangement of simple primitives \cite{Kaick2015ShapeAnalysis, Zhou2015GeneralizedDecomposition} or semantic components \cite{Au2008SkeletonContraction, Berretti20093DGraphs}. Applications of shape decomposition include disciplines such as object recognition and retrieval \cite{Zuckerberger2002PolyhedralApplications, Siddiqi2008RetrievingSurfaces}, shape reconstruction \cite{Goyal2012TowardsModeling}, shape clustering \cite{Averkiou2014ShapeSynth:Synthesis}, or modeling \cite{Funkhouser2004ModelingExample}.

Our motivation for studying shape decomposition comes from biomedical image segmentation. Advanced biomedical imaging techniques, such as 3D electron microscopy, generate large image volumes whose size can range from a gigabyte to hundreds of terabytes \cite{Rubin2014CTAdvance, Abdollahzadeh2019AutomatedMatter, Zheng2018AMelanogaster}. Segmentation of such image volumes generally favors bottom-up strategies, where the image is first over-segmented into supervoxels, then supervoxels are merged subsequently \cite{Lucchi2012, Nunez-Iglesias2014Graph-basedNeuroimages, Funke2019LargeReconstruction}. This strategy is error-prone because both the over-segmentation and subsequent merge are subjected to greedy optimization as opposed to optimizing a global objective. Our idea is instead to approach the segmentation problem based on a top-down strategy, where under-segmentation is followed by subsequent split using \textit{a priori} knowledge of objects to be segmented; in our case, tubularity of neuronal processes or blood vessels. This strategy divides a large image volume into sub-domains whose geometry/topology can be analyzed based on a global objective, independently and in parallel, but necessitates the development of a fast and robust shape decomposition technique capable of processing thousands of tubular structures in a reasonable time.  

This paper develops a novel decomposition algorithm called cylindrical shape decomposition (CSD), decomposing big voxel-based tubular objects in large image volumes. We demonstrate the CSD's application in segmenting tubular structures, as the split operation of a top-down strategy, and its application in decomposing general synthetic objects.

\Figure[t][width=\textwidth]{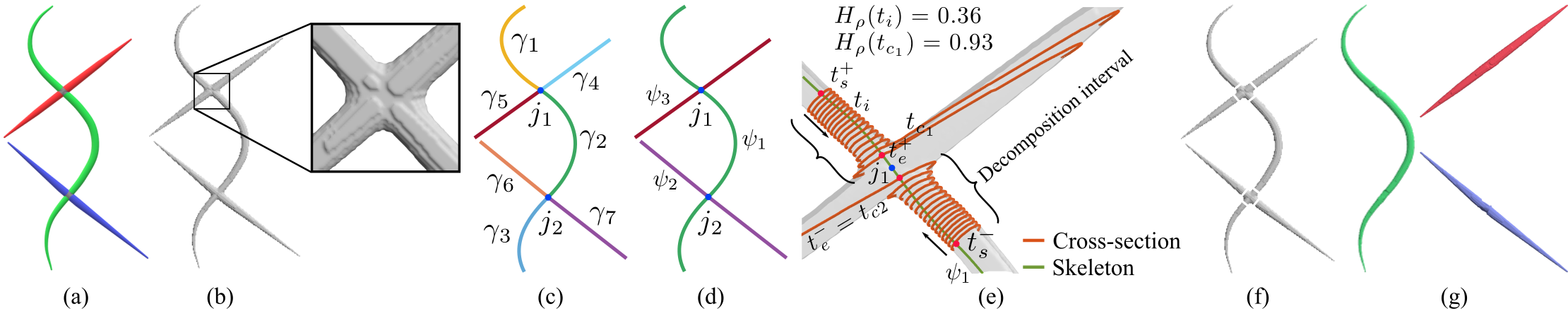}
{Outline of the CSD algorithm. (\textbf{a}) An object is a union of several tubular components. The tubular components are color-coded. (\textbf{b}) A $800 \times 400 \times 70$ voxel-based representation of the object. Intersections of the tubular components are magnified. (\textbf{c}) The curve skeleton of the synthetic object in (\textbf{b}) is the union of all skeleton branches. Skeleton branches are color-coded and denoted as $\gamma$. We define a junction-point $j$ as such a point that skeleton branches connect. Junction-points are shown as blue filled-circles. (\textbf{d}) The curve skeleton of the object is partitioned into maximal-length sub-skeletons $\psi$ over a local orientation cost. The sub-skeletons are color-coded. (\textbf{e}) On a sub-skeleton $\psi$ and in the proximity of a junction-point $j \in \psi$, we define two decomposition intervals. The boundaries of decomposition intervals are shown with red filled-circles. The object is swept along $\psi$ and towards the joint $j$ to find a critical point in each interval. At a critical point, the normalized Hausdorff distance $H_\rho$ between a cross-sectional contour and the mean of visited cross-sectional contours exceeds $\theta_H$. Sweeping directions are shown with arrows. (\textbf{f}) We cut the object at critical points to obtain object parts. (\textbf{g}) The object parts along the same sub-skeleton are assigned the same label to construct a semantic-component. The semantic-components are further reconstructed between their comprising object-parts using generalized cylinders. The synthetic object in (\textbf{a}) comprises seven object-parts, and our algorithm decomposes it into three semantic components. \label{fig:outline}}

\subsection{Related work}
We categorize shape decomposition techniques in the literature into three categories: 1) representing an object in terms of geometrically homogeneous and simple primitives, such as ellipsoids, convex components, or generalized cylinders \cite{Simari2005ExtractionData, RAAB2004VirtualModels, Mortal2004Plumber:Bodies, Asafi2013WeakLines-of-sight, Kaick2015ShapeAnalysis, Zhou2015GeneralizedDecomposition, Li2001DecomposingApplications, Goyal2012TowardsModeling}; 2) decomposing an object into its semantic components using object skeleton or Reeb graph \cite{Reniers2008ComputingMeasure, Au2008SkeletonContraction, Berretti20093DGraphs, Livesu2017ExplicitShapes}; and 3) learning-based decomposition methods \cite{Kalogerakis2010LearningLabeling, Qi2017PointNet:Segmentation, Yu2019Partnet:Segmentation, Deng2020CVXNet:Decomposition}.

Primitives are homogeneous components with a compact representation and efficient computation. Examples of primitives with a simple parametric representation include ellipsoids \cite{Simari2005ExtractionData} and straight cylinders \cite{RAAB2004VirtualModels}. This class of primitives with a simple parametric representation is typically applied in description simplification of complex geometrical models. Therefore, higher-level primitives such as tubular primitives \cite{Mortal2004Plumber:Bodies}, convex components \cite {Asafi2013WeakLines-of-sight, Kaick2015ShapeAnalysis}, generalized cylinders \cite{Zhou2015GeneralizedDecomposition}, and generalized sweep components \cite{Li2001DecomposingApplications, Goyal2012TowardsModeling} were proposed for trading-off the representation simplicity for the generality. For example, tubular primitives in Plumber \cite{Mortal2004Plumber:Bodies} are constructed applying a seeded region growing with a heuristic set of sphere positions and radii. The Plumber does not return a complete decomposition of objects but extracts ideal tubular components \cite{Mortal2004Plumber:Bodies}. Convexity-based methods are another interesting class of high-level primitive-based decomposition techniques, developed based on the human tendency to divide an object into parts around concave regions \cite{Biederman1917Recognition-by-Components:Perception}. An exact decomposition of a shape into convex components is costly and too strict for decomposition because such methods can generate many small parts. Therefore, \cite{Asafi2013WeakLines-of-sight} and \cite{Kaick2015ShapeAnalysis} apply weakly convex components, which are derived from analyzing the pairs of points in the shape visible to each other, obtaining an approximate convex decomposition (ACD). An alternative to the convex decomposition is generalized cylinder decomposition (GCD), quantifying cylindricity. \cite{Zhou2015GeneralizedDecomposition} introduces a quantitative measure for the cylindricity following the minimum description length principle \cite{Grunwald:2007:MDL:1213810} as a measure of the skeleton straightness and the variation among the profiles. In this method, the global objective for merging local generalized cylinders is to minimize the cylindricity. The approximate convexes and generalized cylinders are effective high-level primitives, where the approximate convex method generates smoother cuts between parts, and the generalized cylinders suit better the decomposition of tubular objects. The generalized cylinders method is a computationally expensive technique, e.g., \cite{Livesu2017ExplicitShapes} reported approximately 30 minutes for a single femur decomposition. Another set of high-level primitive-based decomposition techniques is based on cross-sectional sweeping. These methods are computationally less demanding as compared to the convexity and generalized cylinder methods. The sweeping algorithms analyze object cross-sections and generate homogeneous sweeping components. For example, in \cite{Li2001DecomposingApplications}, the object is swept along its curve skeleton in search of critical points, where the object geometry/topology changes substantially. This method uses the variation of the perimeter of consecutive cross-sections as the homogeneity measure, which is sensitive to the surface noise and prone to over-segmentation. The method in \cite{Goyal2012TowardsModeling} generates local prominent cross-sections from a set of initial seed points. This method is semi-automated, requiring user interactions to adjust the density of cross-sections in different object regions and avoid creating prominent cross-sections in regions with no sweep evidence.

To decompose an object into its semantic components, the object curve skeleton or Reeb graph can be used. Both concepts are object descriptors able to guide a decomposition: the curve skeleton is a 1D representation of a 3D object \cite{Cornea2007Curve-skeletonAlgorithms}, encoding its topology and geometry; the Reeb graph tracks topology changes in level sets of a scalar function \cite{Cornea2007Curve-skeletonAlgorithms}. \cite{Reniers2008ComputingMeasure} extracts object curve skeleton based on a collapse measure, i.e., a measure of importance, and subsequently provide an object decomposition by defining skeleton-to-surface mapping based on the shortest geodesics. \cite{Au2008SkeletonContraction} extracts the curve skeleton, applying an implicit Laplacian smoothing with global positional constraints, preserving the mesh connectivity and its key features. \cite{Au2008SkeletonContraction} provides an object decomposition with an approximate measure of thickness about extracted curve skeletons. The tubular decomposition in \cite{Livesu2017ExplicitShapes} aims to be as close as possible to a Voronoi partitioning, having skeleton branches as sites, while satisfying structural constraints that ensure each decomposition element is a tube-like shape. These skeleton-based decomposition methods \cite{Reniers2008ComputingMeasure, Au2008SkeletonContraction, Livesu2017ExplicitShapes} are applied to segment synthetic objects into functional parts, which is not the case for decomposing an under-segmentation error in objects acquired from biomedical imaging datasets. In \cite{Berretti20093DGraphs}, the decomposition of a 3D mesh is a two-step approach accounting for the Reeb graph construction and refinement: the Reeb graph captures the surface topology and protrusions, and the refinement step uses curvature and adjacency information on the graph critical points for fine localization of part boundaries. This approach does not provide smooth boundary cuts between parts, requiring the internal energy function to control the smoothness of boundaries. In the context of skeletonization, it is worth reviewing the L1-medial skeletonization \cite{Huang2013LCloud} and rotational symmetry axis (ROSA) \cite{Tagliasacchi2009CurveCloud} techniques. The L1-medial skeletonization employs localized L1-medians to construct a skeleton. This method uses a weighting function with a supporting radius that defines the size of the local neighborhood; gradually increasing the supporting radius yields a clean and well-connected skeleton. ROSA defines a curve skeleton as a generalized rotational symmetry axis of a shape. The position of a skeleton point in a local set of points is computed by minimizing the sum of the projected distances to the normal extensions of the data points. The L1-medial and ROSA skeletonization techniques may form cycles when two object parts are close to each other, where the skeleton is to be acyclic. These two skeletonization methods deal with incomplete point clouds, but the skeleton centeredness within the objects is not guaranteed. 

Learning-based methods are an important class of shape decomposition techniques, from early statistical modeling methods \cite{Kalogerakis2010LearningLabeling} to recent deep neural network techniques \cite{Qi2017PointNet:Segmentation, Yu2019Partnet:Segmentation, Deng2020CVXNet:Decomposition}. The objective function of a learning-based method is learned from a collection of labeled training objects. Learning-based decomposition methods have demonstrated impressive results, often producing segmentation and labeling comparable to those produced by humans. However, these techniques crucially depend on large training datasets and are often impaired when the objects to be decomposed deviate substantially from the training material.

\subsection{Outline of the CSD algorithm and contributions}
The main idea of the CSD algorithm is to guide the decomposition using the object curve skeleton and cut the object by restricted translational sweeps. Fig. \ref{fig:outline}a shows an under-segmented tubular object as a union of three tubular components. CSD begins with extracting the object curve skeleton (Fig. \ref{fig:outline}c) and partitioning the skeleton into maximal-length sub-skeletons over an orientation cost function (Fig. \ref{fig:outline}d). Each sub-skeleton corresponds to a semantic tubular component. To identify intersections of the semantic components, CSD translationally sweeps the object along sub-skeletons, searching for critical points where the object cross-section changes substantially (Fig. \ref{fig:outline}e). A translational sweep is restricted in decomposition intervals; in the proximity of junction-points where sub-skeletons intersect. The object is cut at critical points to obtain object parts (Fig. \ref{fig:outline}f). A semantic component is further reconstructed at intersections, using generalized cylinders (Fig. \ref{fig:outline}g).

The CSD algorithm possesses several advantages over previous shape decomposition techniques. Unlike semi-automated methods in \cite{Goyal2012TowardsModeling} and \cite{Au2008SkeletonContraction}, CSD is a fully automatic algorithm and requires no manual interventions; the decomposition is guided using an algorithm that partitions the object skeleton curve into distinct maximal-length straight sub-skeletons. Compared to primitive-based methods in \cite{Simari2005ExtractionData, RAAB2004VirtualModels, Kaick2015ShapeAnalysis, Zhou2015GeneralizedDecomposition}, or skeleton-to-surface mapping in \cite{Reniers2008ComputingMeasure, Berretti20093DGraphs}, our method identifies the intersection of the object parts and defines smooth boundary cuts between them. Compared to \cite{Li2001DecomposingApplications}, our method is intrinsically more robust in defining critical points in the presence of noise because we measure cross-sectional changes using the mean close curve of traversed cross-sections and modified Hausdorff distance. In \cite{Zhou2015GeneralizedDecomposition}, generating a generalized cylinder requires iterative operations, yet many such primitives are required to cover the object; henceforth, these primitives must be merged to satisfy a global objective. Such methods are computationally expensive, not suitable for the decomposition of big voxel-based objects in large image volumes. Unlike \cite{Reniers2008ComputingMeasure, Au2008SkeletonContraction, Livesu2017ExplicitShapes}, and \cite{Berretti20093DGraphs}, which apply a one-to-one assignment between skeleton branches and object parts, we propose to merge skeleton branches belonging to the same semantic part. Unlike learning-based techniques \cite{Kalogerakis2010LearningLabeling, Qi2017PointNet:Segmentation, Yu2019Partnet:Segmentation, Deng2020CVXNet:Decomposition}, our proposal does not rely on training and thus generalizes to the variation of tubular objects extracted from medical images, where it attains consistent quality without the need for additional training datasets. In comparison to the L1-medial and ROSA skeletonization techniques, we use a distance-based skeletonization approach, which correctly stays at the center of the object, even when the object parts are adjacent.

In the experimental section of this paper, we demonstrate the application of CSD in the segmentation of large electron microscopy volumes of myelinated axons. We also demonstrate the CSD decomposition of vascular networks and synthetic objects. Moreover, we compare CSD to other state-of-the-art decomposition techniques (ACD \cite {Kaick2015ShapeAnalysis} and GCD \cite{Zhou2015GeneralizedDecomposition}), and our skeletonization technique to well-known skeletonization approaches (L1-medial and ROSA). We also evaluate the effect of surface noise on decomposition results and assess a methodology to reduce the CSD computation-time.

\section{Preliminaries}
This section defines the core concepts used in the paper as there are no generally accepted definitions for most of them.

\textbf{Object.} An object $\Omega \subset \mathbb{R}^3$ is a nonempty bounded open set. We assume that its boundary $\partial \Omega$ is homeomorphic to a 2-sphere. For a discrete object, which results from foreground segmentation, we define a 3D binary image as $I: X \subset \mathbb{Z} ^3 \to \{0,1\}$, and a segmented object $\Omega := \{x \in X: I(x) = 1 \}$, where $X$ is the image domain. Throughout the paper, $\Omega$, $\partial \Omega$, and $x$ are in $\mathbb{R}^3$ unless defined otherwise.

\textbf{Curve skeleton.} Given $\Omega$ and $\partial \Omega$, the curve skeleton $\Upsilon \subset \Omega$ is defined as a locus of centers of maximal inscribed balls \cite{Lieutier2004AnyAxis}. A ball $B(x,r)$ centered at $x \in \Omega$ with radius $r$ is maximally inscribed if its  surface touches $\partial \Omega$ in at least two distinct points. Formally, $B$ is a maximal inscribed ball in $\Omega$ if $\forall B',\; B \subseteq B' \subseteq \Omega \Rightarrow B' = B$. 

\textbf{Curve skeleton point type.} We distinguish three types of points on the curve skeleton of an object: 1) regular-points that have exactly two neighbor points on the skeleton, 2) end-points that have exactly one neighbor point on the skeleton, and 3) junction-points that have three or more neighbor points on the skeleton \cite{Cornea2007Curve-skeletonAlgorithms}. We denote the collection of junction-points as $J$ where $j \in J$ and the collection of end-points as $O$ where $o \in O$.

\textbf{Skeleton branch.} Removing junction-points $J$ from the curve skeleton $\Upsilon$ results in disconnected simple curves, known as skeleton branches. The collection of skeleton branches is denoted as $\Gamma$, and a skeleton branch is $\gamma \in \Gamma$. For $\gamma (t): [0,1] \to \mathbb{R}^3$, its arc-length is written as $ l = \int_{0}^{1} |{\dot \gamma(t)}| \, \mathrm{d}t $ with the convention $\dot \gamma(t) := \dfrac{\mathrm{d}}{\mathrm{d}t}\gamma(t)$. 

\textbf{Skeleton graph.} The topology of curve-skeleton $\Upsilon$ can be represented as a connected acyclic undirected graph (i.e., a tree) $G_{\Upsilon} = (V, E, L)$. There is a one-to-one map between skeleton branches in $\Gamma$ and edges in $E$ and a one-to-one map between the union of end-points and junction-points ($O \cup J$) and vertices in $V$. This means that for each branch $\gamma \in \Gamma$ we associate exactly one edge in $e$ in $G_{\Upsilon}$. $L \subset \mathbb{R}^+$ is the set of edge lengths. The length of an edge is the arc-length of its associated skeleton branch.

\textbf{Walk, path.} A walk is a finite or infinite sequence of edges which joins a sequence of vertices. A finite walk is a sequence of edges $W = \{e_1, e_2, \ldots, e_{n'}\}$ for which there is a sequence of vertices $\{v_0, v_1, \ldots, v_{n'}\}$ such that $e_i = v_{i-1}v_{i}$ for $i = 1, \dots , n'$. The vertex sequence of the walk is $(v_0, v_1, \ldots, v_{n'})$. A path is a walk in which all vertices are distinct.

\textbf{Sub-skeleton} is a path in the curve skeleton domain. If $W = \{e_1, e_2, \ldots, e_{n'}\}$ is a path in the skeleton graph, and $\{\gamma_1, \gamma_2, \ldots, \gamma_{n'}\}$ are corresponding skeleton branches, then $\psi = \cup_i \gamma_i \subseteq \Upsilon$ is a sub-skeleton.

\textbf{Critical point.} A point on a sub-skeleton at which the cross-sectional contour of the object changes substantially. We provide a formal definition in section \ref{sec:critical_point}.

\textbf{Cut.} A closed simple curve $C \subset \partial \Omega$ is a cutting-curve if $\partial \Omega \setminus C$ is not connected. Cut means removal of a cutting-curve from the surface.

\section{Outline of the CSD algorithm}
The outline of the CSD algorithm is shown in Fig. \ref{fig:outline}, and it is as follows: 
\begin{enumerate}
    \item define the curve skeleton of a given object (Fig. \ref{fig:outline}c, section \ref{sec:skel});
    \item partition the curve skeleton of the object into sub-skeletons (Fig. \ref{fig:outline}d, section \ref{sec:skel_partition});
    \item define decomposition intervals to restrict the object sweep (Fig. \ref{fig:outline}e, section \ref{sec:dec_int});
    \item sweep the object to find critical points and cut the object at critical points (Fig. \ref{fig:outline}e and Fig. \ref{fig:outline}f, section \ref{sec:critical_point});
    \item reconstruct the object between parts that have the same label using generalized cylinders (Fig. \ref{fig:outline}g, section \ref{sec:obj_rec}).
\end{enumerate}
The CSD algorithm is designed for genus zeros objects.

\section{Skeleton partitioning} \label{sec:skel_part}
We use the curve skeleton of an object to drive the decomposition. For that, we partition the skeleton graph into several distinct paths union of which covers the skeleton graph. The partitioning of the skeleton graph, by extension, partitions the curve skeleton into sub-skeletons. Each sub-skeleton corresponds to exactly one semantic object component.

\subsection{Curve skeleton} \label{sec:skel}
To determine the curve skeleton of an object $\Omega$ with sub-voxel precision, we apply a method from \cite{Hassouna2005RobustSets} and \cite{VanUitert2007SubvoxelMethods}. The algorithm initiates by determining a point $x^* \in \Omega$ with the biggest distance from the object surface $\partial \Omega$ inside the object domain. This point is used to determine a skeleton branch $\gamma(t): [0,1] \to \mathbb{R}^3$, starting at $x_s$, the furthest geodesic point from $x^*$ in $\Omega$, and ending at $x^*$. A cost function $F$ is defined to enforce the path to run in the middle of $\Omega$, where $F$ should increase if the path moves away from the center. To determine $F$, we find the distance field $D(x)$ from $\partial \Omega$, and assign $F = 1 - \big(\dfrac{D(x)}{D(x^*)}\big)^2$. The distance field $D(x)$ is determined by solving an Eikonal equation on the object domain $\Omega$ using the fast marching method \cite{Sethian1996AFronts}. Starting at $x_s$, the skeleton branch $\gamma$ is traced by a back-tracking procedure on $F$ to reach $x^*$, written as
\begin{equation}
    \gamma = \displaystyle \arg \min_{P} {\displaystyle \int_{x_{s}}^{x^*}}
     F (P(t)) \, \mathrm{d}t,
\label{eq:skeleton_path}
\end{equation}
where $t$ traces the path $P$. We use the Euler scheme for the back-tracking procedure, which solves the ordinary differential equation with a sub-voxel accuracy. This process is repeated to determine further branches that form the curve skeleton of the object. But rather than using the single point $x^*$ as the staring point for the fast marching method, all points in the previously calculated branches are used as starting points. We propagate a new wave from the starting points with the speed $F$ to update $x_s$. The point $x_s$ is now the furthest point from the current state of the curve skeleton and the starting point of the new branch. Applying a back-tracking algorithm from the updated $x_s$ defines the new skeleton branch. Fig. \ref{fig:skeleton} shows the skeletons of two synthetic objects and a vascular network.

\Figure[t][width=0.99\columnwidth]{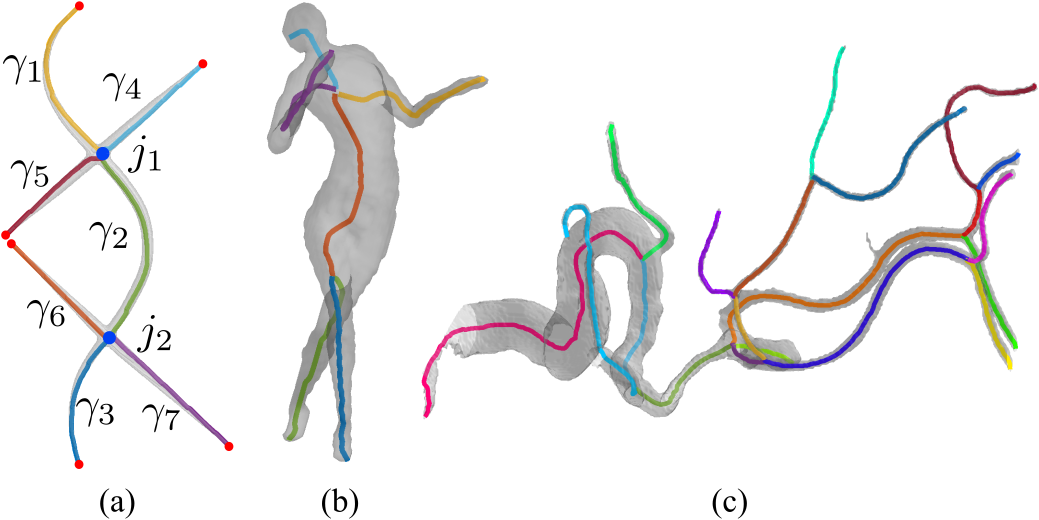}
{The curve skeleton of an object is the union of all skeleton branches. (\textbf{a}) The skeleton of the synthetic object, size: $800 \times 400 \times 70$ voxels, seven branches. The blue filled-circles show junction-points, and the red filled-circles show end-points. The skeleton graph of this object is $G_\Upsilon(V,E,L)$, where $E = \{e_1,\ldots, e_7\}$, $V=\{v_0, \ldots, v_7 \}$, and $L =\{l_1, \ldots, l_7 \} $. (\textbf{b}) The skeleton of a synthetic object, size: $128 \times 128 \times 128$ voxels, six branches. (\textbf{c}) The skeleton of a vascular network, size: $256\times256\times256$ voxels, 20 branches. Skeleton branches are color-coded. \label{fig:skeleton}}

\subsection{Skeleton graph decomposition} \label{sec:skel_partition}
Several skeleton branches are often required to represent one semantic component of an object, and therefore detecting skeleton branches is not sufficient for a semantic decomposition. An example is shown in Fig. \ref{fig:skeleton}a, where the union of three skeleton branches $\gamma_1$, $\gamma_2$, and $\gamma_3$ is required to represent one tubular component. To formalize what constitutes a semantic decomposition, we consider connectivity, length, and local orientation, to unify skeleton branches. We propose an algorithm for traversing the graph representation of the curve skeleton $G_\Upsilon (V,E,L)$, decomposing $G_\Upsilon$ into distinct paths, each corresponds to a semantic component. The algorithm starts at the root edge and explores as far as possible along edges, which provide the optimal choice at each stage.

\Figure[t][width=0.99\columnwidth]{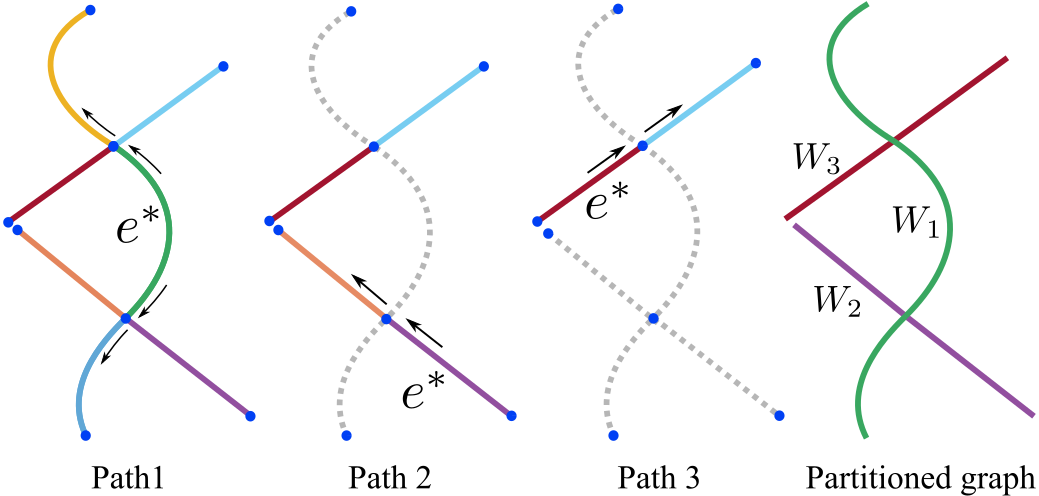}
{Partitioning the skeleton graph of the synthetic tubular object. $E$ comprises seven edges and is partitioned into three paths: $W_1 = \{ e_1, e_2, e_3 \}$, $W_2 = \{ e_4, e_5 \} $, and $W_3 = \{ e_6, e_7 \}$. We determine $W_1$ starting from the longest edge in $E$ denoted as $e^*$ towards its incident vertices. At each vertex, we traverse the edge with minimum orientation cost. Appending new edges terminates when a leaf vertex is visited, or the angle between two successive edges is smaller than $\theta_c$. We subtract $W_1$ from $E$, when $W_1$ is determined (see Algorithm \ref{alg:walk}). The blue filled-circles show vertices in $G_\Upsilon$. The edges are color-coded with full-lines. At vertices, arrows show where to traverse next when standing on $e^*$. The grey dash-lines show the previously calculated paths. \label{fig:skel_partition}}

We partition the skeleton graph of the object into several distinct paths union of which covers the set of graph edges. Formally, we partition $G_\Upsilon (V,E,L)$ into $m$ paths $W_i, i = 1, \dots ,m$ so that $\cup_i W_i = E$ and $W_i \cap W_k = \emptyset \; \forall i,k = 1, \dots, m,\; i \neq k$. To determine the paths, we require four conditions: 1) the path contains the longest edge not associated to any other path, 2) the path has the maximum number of edges, 3) the associated angle between two successive edges is bigger than $\theta_c$, and 4) the path locally minimize an orientations cost. Denoting two successive edges in a path as $e_s$ and $e_{s+1}$, the edge $e_{s+1}$ has the maximum angle compared to $e_s$ among the set of connected edges to $e_s$. The angle between two edges $e_s$ and $e_{s+1}$ is the angle between the line segments connecting endpoints of the skeleton branches associated with edge $e_s$ and $e_{s+1}$, and it lies in range $[0, \pi]$. We used Algorithm \ref{alg:walk} to determine the $m$ distinct paths on $G_\Upsilon$. Fig. \ref{fig:skel_partition} shows skeleton graph decomposition of the synthetic object $n = 7$ into three paths $m = 3$. Each path is equivalent to a sub-skeleton.

\begin{algorithm}[t]
\DontPrintSemicolon
\SetKwInOut{Input}{Input}\SetKwInOut{Output}{Output}
\Input{$G_\Upsilon=(V,E,L)$; $\theta_c$.}
\Output{Collection of distinct paths $\Lambda$.}
    \SetAlgoLined
    $\Lambda \leftarrow \emptyset$ \;
    \While {$E \neq \emptyset$}{
    $W \leftarrow \emptyset$; $e^* \leftarrow $ longest $e \in E$ \;
    $V^* \leftarrow $ \{ incident vertices to $e^*$ \} \;
    $W \leftarrow W \cup \{ e^* \}$ \;
    \ForAll {$\upsilon \in V^*$}{
        $e^{ref} \leftarrow e^*$ \;
        $\upsilon^{next} \leftarrow \upsilon$ \;
        \While {$ deg(\upsilon^{next}) > 1 $ {\bf and} $e^{ref} \neq \emptyset$}{ 
            $CE \leftarrow$ \{  edges connected to $\upsilon^{next}$ \} $\setminus e^{ref}$ \;
            $e^{next} \leftarrow \emptyset$ \;
            \ForAll {$e^{ngb} \in CE$}{
                $\theta_{max} \leftarrow \theta_c$; $\theta \leftarrow \angle (e^{ref}, e^{ngb})$ \;
                \If {$\theta > \theta_{max}$}{
                    $\theta_{max} \leftarrow \theta$; $e^{next} \leftarrow e^{ngb}$ \;
                    }
                    }
            $e^{ref} \leftarrow e^{next}$ \;
            \If{$e^{ref} \neq \emptyset$}{
                $\upsilon^{next} \leftarrow \upsilon_2$, where $e^{ref} = (\upsilon_2, \upsilon^{next})$ \;
                $W \leftarrow W \cup \{ e^{ref} \}$ \;
            }
        }
    }
    $\Lambda \leftarrow \Lambda \cup \{W\}$ and $E \leftarrow E \setminus W$ \;
    }
    \caption{Decomposing the set of edges of $G_\Upsilon$ into distinct paths. A vertex and an edge are called incident if the vertex is one of the two vertices the edge connects.}
    \label{alg:walk}
\end{algorithm}

\section{Cylindrical decomposition} \label{sec:critical_p}
In this section, we propose a method to decompose an object into parts and intersections by cutting the object at critical points. To determine critical points, we sweep the object along sub-skeletons in decomposition intervals to find locations where the object geometry changes substantially (see Fig. \ref{fig:cross_section}).

\subsection{Decomposition interval} \label{sec:dec_int}
We restrict the sweep of the object along each sub-skeleton to decomposition intervals in the proximity of a junction-point $j$ on sub-skeleton $\psi$, as illustrated in Fig. \ref{fig:dec_interval}. It is convenient to work with parametrized sub-skeleton $\psi (t): [0,1] \to \mathbb{R}^3$. We define two decomposition intervals  $[t_s^+, t_e^+]$ and $[t_e^-, t_s^-]$ for each junction-point as in Fig. \ref{fig:dec_interval}a. To determine the boundaries of a decomposition interval, we define an upper threshold $r_s$ and a lower threshold $r_e$. We specify $r_s$ and $r_e$ based on the radius of the maximal inscribed ball at $t_j$ and two factors $\alpha_s \geq 1$ and $\alpha_e \geq 0$ where $\alpha_s \geq \alpha_e$, as $r_s = \alpha_s \times r$ and $r_e = \alpha_e \times r$.

To determine the thresholds, we use the signed arc-length from $j$. Define $t_j$ so that $j = \psi(t_j)$. Then $t_s^+$ ($t_s^-$) is such a point on the sub-skeleton that signed arc-length from $t_j$ to $t_s^+$ ($t_s^-$) equals $r_s$ ($-r_s$). And $t_e^+$ ($t_e^-$) is such a point on the sub-skeleton that signed arc-length from $t_j$ to $t_e^+$ ($t_e^-$) equals $r_e$ ($-r_e$). We have $t_s^+ < t_e^+ < t_j < t_e^- < t_s^-$.  

The upper and lower thresholds may imply arc-distances outside parametrization limits of $\psi$. If the arch-length from $\psi(0)$ to $\psi(t_j)$ is smaller than $r_s$ ($r_e$) we assign $t_s^+ = 0$ ($t_e^+ = 0$). And if the arch-length from $\psi(t_j)$ to $\psi(1)$ is smaller than $r_s$ ($r_e$) we assign $t_s^- = 1$ ($t_e^- = 1$). Also, when a junction-point is at the either ends of a sub-skeleton, e.g., in a T-shape object, we define only one decomposition interval. Therefore, if $\psi(t_j) = 0$ ($\psi(t_j) = 1$) the only interval that we define is $[t_e^-, t_s^-]$ ($[t_s^+, t_e^-]$). Fig. \ref{fig:dec_interval}b shows decomposition intervals in the proximity of $j_1$ and $j_2$ on sub-skeletons $\psi_1, \psi_2$, and $\psi_3$.

\Figure[t][width=0.99\columnwidth]{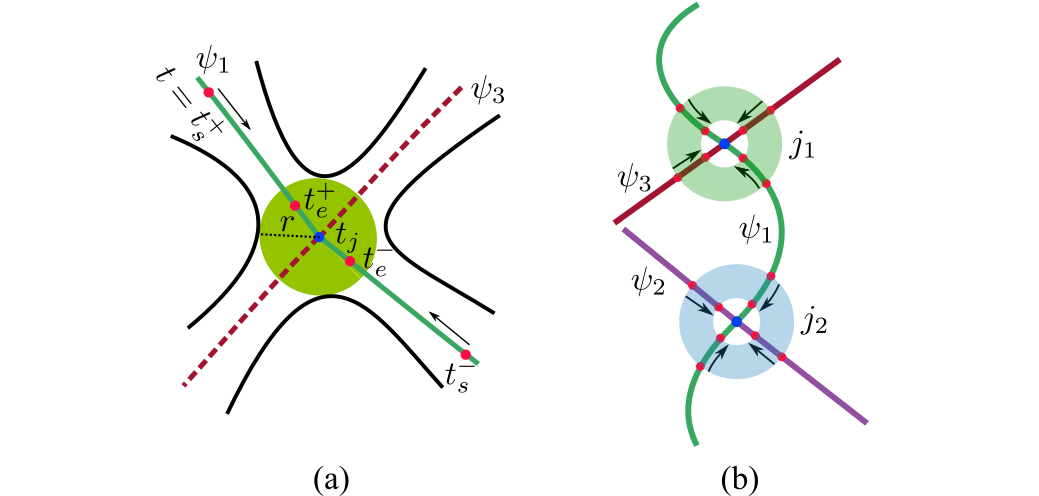}
{(\textbf{a}) In the proximity of every junction-point, e.g. $j_1$ blue filled-circle, and on each sub-skeleton, e.g. $\psi_1$ green line, we define two decomposition intervals, $[t_s^+, t_e^+]$ and $[t_e^-, t_s^-]$, tracing $\psi_1$, from $t_s^+$ to $t_e^+$ and from $t_s^-$ to $t_e^-$ (red filled-circles). The lower and upper bounds of the intervals are two factors of the radius of the maximal inscribed ball at $t_j$, the green circle. (\textbf{b}) Decomposition intervals in the proximity of all junction-points $j_1$ and $j_2$ and for all sub-skeletons $\psi_1$, $\psi_2$, and $\psi_3$ are defined with the red filled-circles. Only in decomposition intervals, we are allowed to sweep the object. Arrows depict the sweeping direction to approach junction-points. \label{fig:dec_interval}}

\subsection{Critical point} \label{sec:critical_point}
A critical point on a sub-skeleton is such a point that the cross-sectional contour of the object at this point changes substantially (Fig. \ref{fig:cross_section}). We use the Hausdorff metric to compare geometrical changes between cross-sectional contours in a decomposition interval. The Hausdorff distance between two curves $C_1$ and $C_2$ is calculated as
\begin{align}
	\displaystyle \mathcal{H}(C_1, C_2) &= \nonumber\\
	\max &\{\sup_{p \in C_1} \; \inf_{q \in C_2} \; d(p, q),\; \sup_{q \in C_2} \; \inf_{p \in C_1} \; d(p, q)\},
\label{eq:HausD}
\end{align}
where d(.) is the Euclidean distance between two points. We sweep $\partial \Omega$ by a cross-sectional plane $\mathcal{P} \subset \mathbb{R}^3$ to extract the cross-sectional contours. A cross-sectional plane $\mathcal{P}(t)$ is a plane orthogonal to $\psi$ at every point $t$ along $\psi$. The plane normal is equal to the tangent vector to $\psi$ at point $\psi(t)$. We sweep $\partial \Omega$ by $\mathcal{P}$ along $\psi$ in $[t_s^+,t_e^+]$ interval starting at $t_s^+$ toward $t_e^+$, and in $[t_e^-,t_s^-]$ interval starting at $t_s^-$ toward $t_e^-$, as illustrated in Fig. \ref{fig:dec_interval}. Let $\mathcal{P}(t)$ intersects $\partial \Omega$ at an inquiry point $t$. Since we assumed that $\partial \Omega$ is homeomorphic to a 2-sphere, the cross-sectional contour $C(\varsigma):[0,1] \to \mathbb{R} ^2$ is a simple closed curve, where $C(0) = C(1)$. Translating $\mathcal{P}$ along $\psi(t)$ with $t$ moving in decomposition intervals, we compare the Hausdorff distance between the cross-sectional contour at $t$ denoted as $C_t$ with the average of visited cross-sectional contours $\mu$.

\Figure[t][width=0.99\columnwidth]{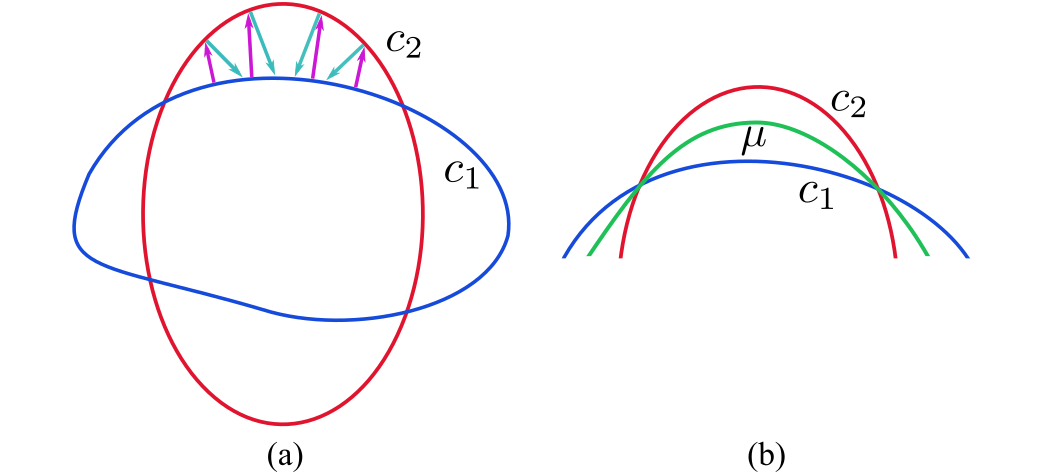}
{(\textbf{a}) Two nearly similar curves $C_1$ and $C_2$. Turquoise arrows represent $OM(C_1,C_2)$, and pink arrows represent $OM(C_2,C_1)$. (\textbf{b}) The average curve $\mu$ obtained from the orthogonal correspondence between $C_1$, an already visited curve, and $C_2$ a new cross-sectional curve. \label{fig:mean_curve}}

To find the average curve $\mu$ between two nearly similar curves $C_1$ and $C_2$, we first need a one-to-one orthogonal mapping (OM) between $C_1$ and $C_2$. Consider that $C_1$ is parameterized by $\varsigma$. To each point $C_1(\varsigma)$ of $C_1$, the $OM(C_1,C_2)$ associates the closest point $C_2(\varsigma)$ on $C_2$ that lies on the line passing through $C_1(\varsigma)$ and having for direction the normal $N(\varsigma)$ to $C_1$ at $C_1(\varsigma)$. Having this mapping, then each point $C_2(\varsigma)$ of $C_2$ may be expressed as the normal offset $C_1(\varsigma) + d(\varsigma)N(\varsigma)$ of $C_1(\varsigma)$. We say that $C_1(\varsigma)$ is the closest normal projection of $C_2(\varsigma)$ onto $C_1$ and can express $C_2$ as a deformation of $C_1$ completely defined by the normal displacement field $d(\varsigma)$ (see Fig. \ref{fig:mean_curve}a) \cite{Chazal2005Projection-homeomorphicSurfaces}. The average curve obtained by this orthogonal correspondence is asymmetric: $OM(C_1, C_2)$ is not necessarily equal to $OM(C_2, C_1)$. Therefore, we consistently take $C_1$ as an already visited curve, $C_2$ as the new cross-sectional curve, and define the average curve over $OM(C_1,C_2)$ (see Fig. \ref{fig:mean_curve}).

We normalize the Hausdorff distance $\mathcal{H}(C_t, \mu)$ to the range $[0, 1]$ and denote it as $H_\rho(t)$. For that we first find a point interior to $C_t$ denoted as $\kappa$. We define $\kappa \in \mathbb{R}^2$ to be the intersection of $\mathcal{P}$ and $\psi$ at point $t$. Defining $d_{C_t}(\kappa) = \sup_{q \in C_t} d(\kappa, C_t)$, we write $H_\rho(t)$ as
\begin{equation}
	\displaystyle H_\rho(t) = \dfrac{\mathcal{H}(C_t,\mu)}{\mathcal{H}(C_t,\mu) + d_{C_t}(\kappa)}.
\label{eq:Hous_th}
\end{equation}

We define a similarity threshold between cross-sectional contours as $\theta_H$. While sweeping $\partial \Omega$ along $\psi$ from $t_s^+$ ($t_s^-$) to $t_e^+$ ($t_e^-$), if $H_\rho(t) < \theta_H$, the inquiry continues to the next point. However, if $H_\rho(t) \geq \theta_H$ the inquiry stops at $t$ and the point is called a critical point, denoted as $t_{c_1}$ ($t_{c_2}$), as shown in Fig. \ref{fig:cross_section}. In $[t_s^+, t_e^-]$ ($[t_e^-, t_s^-]$), if at no inquiry point $H_\rho(t)$ exceeds $\theta_H$, we define the $t_{c_1}$ ($t_{c_2}$) as the point with minimum arc-distance $r$ ($-r$) to $\psi(t_j)$ at which $H_\rho$ is maximum.

\Figure[t][width=0.99\columnwidth]{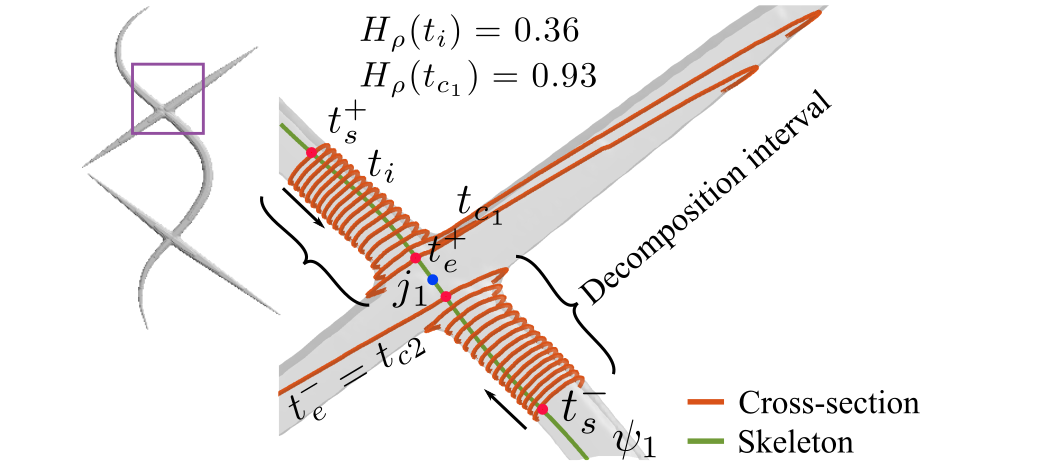}
{Sweeping the object surface along the sub-skeleton $\psi_1$ at junction-point $j_1$ (blue filled-circle) between $[t_s^+,t_e^+]$ and $[t_s^+,t_e^+]$ (red filled-circles). Sweeping directions are shown with arrows. At any decomposition interval, if $H_\rho < \theta_H$ the inquiry continues to the next point. If $H_\rho(t) \geq \theta_H$ the inquiry stops at $t$ and the point is called a critical point. The critical point in the first interval is denoted as $t_{c_1}$ and in the second interval is denoted as $t_{c_2}$. \label{fig:cross_section}}

\Figure[t][width=0.99\columnwidth]{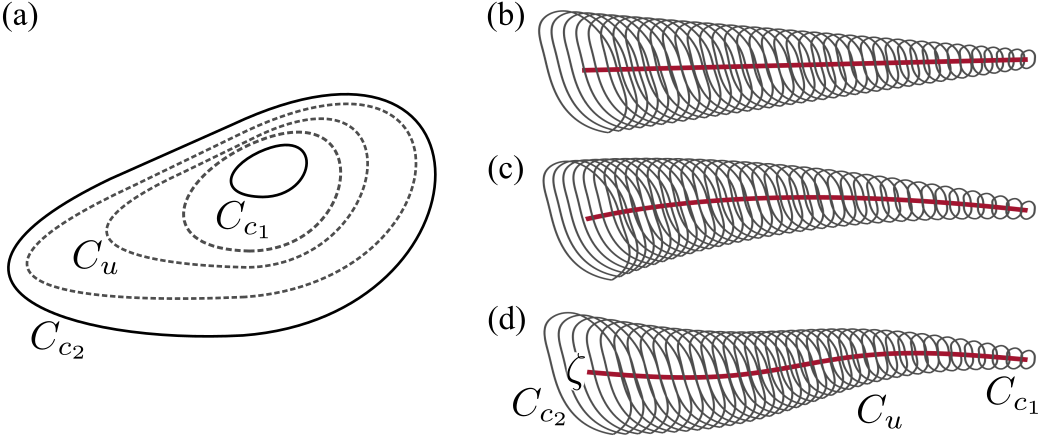}
{(\textbf{a}) Homotopy between two curves $C_{c_1}(\varsigma)$ and $C_{c_2}(\varsigma)$. Generalized cylinder along (\textbf{b}) a linear, (\textbf{c}) spline, and (\textbf{d}) sine interpolation between $\psi(t_{c_1})$ and $\psi(t_{c_2})$. \label{fig:gc}}

\section{Object reconstruction} \label{sec:obj_rec}
We cut the object at all critical points and decompose $\partial \Omega$ into $n$ parts, $n$ is the number of skeleton branches, and $\delta$ intersections, $\delta$ is the number of junction-points. We distinguish between an object part and an intersection such that the interior of an intersection includes a junction-point. The final decomposition step is to discard intersections and assign the same label to those object parts that are along the same sub-skeleton to obtain $m$ semantic tubular components, $m$ is the number of sub-skeletons. As we discard the intersections, we reconstruct the semantic tubular components using generalized cylinders. A generalized cylinder $\Phi(u,\varsigma) : [0,1]^2 \to \mathbb{R}^3$ represents an elongated surface on an arbitrary axis and smoothly varying cross-sections \cite{SHANI1984129}. In Cartesian coordinates $x_1,x_2,x_3$, the axis is parametrized by $u$ as $\zeta(u) = (x_1(u), x_2(u), x_3(u))$ and cross-section boundary is represented as $C_u(\varsigma) = (x_1(u, \varsigma), x_2(u, \varsigma))$. To construct $\Phi$, we apply a translational sweep along $\zeta(u)$ using closed simple curves $C_u(\varsigma)$ written as
\begin{align}
    \Phi(u,\varsigma):=\{\zeta(u) \in \mathbb{R}^{3}, C_u(\varsigma) \in \mathbb{R}^{2} : u,\varsigma \in [0,1] \}.
\label{eq:genCylinder}
\end{align}

To obtain a parametric representation of generalized cylinders, it is convenient to employ a local coordinate system defined with the origin at each point of $\zeta(u)$. A convenient choice is the Frenet-Serret frame which is suitable for describing the kinematic properties of a particle moving along a continuous, differentiable curve in $\mathbb{R}^3$. The Frenet-Serret frame is an orthonormal basis composed of three unit vectors $e_T$, $e_N$, and $e_B$, where $e_T$ is the unit tangent vector, and $e_N$ and $e_B$ are the unit normal and unit binormal vectors, respectively. By defining the cross-section in the Frenet-Serret frame, we form a parametric representation of generalized cylinders \cite{BallardDanaHarryandBrown1982ComputerVision} as follows:

\begin{align}
    \Phi(u,\varsigma) = \zeta(u) + x_1(u, \varsigma) e_N(u) + x_2(u, \varsigma) e_B(u) 
\end{align}

To define $C_u(\varsigma)$, we use a homotopy between two curves $C_{c_1}(\varsigma)$ and $C_{c_2}(\varsigma)$, where the curves are obtained by cross-sectioning the object surface at critical points $t_{c_1}$ and $t_{c_2}$, respectively (see Fig. \ref{fig:gc}a). Let the simple closed curves $C_{c_1}(\varsigma)$ and $C_{c_2}(\varsigma)$ in $\mathbb{R} ^2$ be homotopic with a continuous map $h: [0, 1]^2 \to \mathbb{R} ^2$. So, we write:
\begin{align}
    h(0,\varsigma) &= C_{c_1}(\varsigma), \; h(1,\varsigma) = C_{c_2}(\varsigma), \; \forall \varsigma \in [0,1], \\
    h(u,0) &= h(u,1), \forall u \in [0,1],
\label{eq:homotopy}
\end{align}
where $h$ is called a homotopy from $C_{c_1}(\varsigma)$ to $C_{c_2}(\varsigma)$. We denote a cross-section at a point along $\zeta(u)$ as $C_u := h(u,.)$. Note that, $\mathbb{R} ^2$ is simply connected space. We use a linear homotopy between $C_{c_1}(\varsigma)$ to $C_{c_2}(\varsigma)$ defined as:
\begin{equation}
    h(u,\varsigma) = (1-u) \; C_{c_1}(\varsigma) + u \; C_{c_2}(\varsigma),
\label{eq:linear_homotopy}
\end{equation}
where the computation on the right side is in $\mathbb{R} ^2$. Equation \eqref{eq:linear_homotopy} essentially indicates that we are moving from $C_{c_1}(\varsigma)$ to $C_{c_2}(\varsigma)$ along a straight line. To define the curve $\zeta(u)$, we use an interpolation between $\psi(t_{c_1})$ and $\psi(t_{c_2})$. Figs. \ref{fig:gc}b-d show $\Phi$ on different choices of $\zeta$.

\section{Experimental results}
In this section, we evaluate the effect of CSD parameters on object decomposition, present the applications and advantages of the proposed method in decomposing tubular objects, and show the performance of CSD applied on more general objects. We use the marching cubes algorithm \cite{Lorensen1987MarchingAlgorithm} to compute a triangulated mesh of the object surface, visualizing voxel-based objects. 

\subsection{Parameter setting}
In all our experiments, we fix the value of $\alpha_e = 1$ (section \ref{sec:dec_int}), meaning that the distance of $t_e^+$ ($t_e^-$) to a junction-point is equal to the radius of the maximal inscribed ball at that junction-point. Here, we examine the effect of $\alpha_s$, which determines $t_s^+$ ($t_s^-$) of the decomposition intervals (section \ref{sec:dec_int}) for values equal to 10, 20, and 30. For that, we set the similarity threshold $\theta_H$ (section \ref{sec:critical_point}) equal to 0.7. We use a linear interpolation to define $\zeta$, the curve on which a generalized cylinder is defined (section \ref{sec:obj_rec}), and set the value of the angular threshold $\theta_c$ (section \ref{sec:skel_partition}) equal to $0 ^\circ$. Figs. \ref{fig:alpha_s}a-c show the decomposition of the synthetic tubular object for $\alpha_s$ equals 10, 20, and 30, respectively. The decomposition/reconstruction at $\alpha_s = 10$ is more faithful to the original object as the critical points are detected close to the junctions. Increasing the value of $\alpha_s$ enlarges the decomposition interval. Therefore, at $\alpha_s = 30$, CSD detects the critical points distant from the junctions, resulting in a bigger reconstruction error compared to small values of $\alpha_s$. It is worth noting that although setting $\alpha_s$ to small values provides more accurate decomposition/reconstruction results, it may also result in defining a critical point within an intersection. This can be the case when applying CSD to objects degraded with surface noise. The curve skeleton of an object with surface noise may not exactly lie in the center of the object, which means that the junction-point can be dislocated and the radius of the maximum inscribed ball at that junction-point be measured smaller than its true value. For $\alpha_s$ and $\alpha_e$, we suggest values in range [3, 20] and [0.5, 2], respectively.

We also examine the effect of $\theta_H$ value, which is the similarity threshold between cross-sectional contours and $\mu$. Figs. \ref{fig:theta_H}a-c show the decomposition of the tubular synthetic object at $\theta_H$ equals 0.6, 0.7, and 0.8, respectively. To better demonstrate the effect of $\theta_H$, we set $\alpha_s = 30$, which we earlier showed this could result in a substantial decomposition error. We use a linear interpolation to define $\zeta$ and set $\theta_c = 0 ^\circ$. At $\theta_H = 0.6$, CSD is sensitive to cross-sectional changes and does not tolerate the gradual increase of the tube diameter; hence critical points are detected distant from junction-points, and the reconstruction shows a low agreement with the original object. Increasing the value of $\theta_H$ to 0.7 increases the tolerance of CSD to cross-sectional changes. Therefore, despite distant starting points from junction-points, the reconstruction shows a better agreement to the original object, and at $\theta_H = 0.8$, the reconstructed object is faithful to the original object. Note that increasing $\theta_H$ elevates the tolerance of CSD to the cross-sectional changes quickly, e.g., at $\theta_H = 0.9$, the algorithm tolerates a nine times difference between a cross-section and $\mu$, and at $\theta_H = 0.95$, it tolerates a 19 times difference. We suggest $\theta_H$ to be in the range [0.7, 0.85].

\Figure[t][width=0.99\columnwidth]{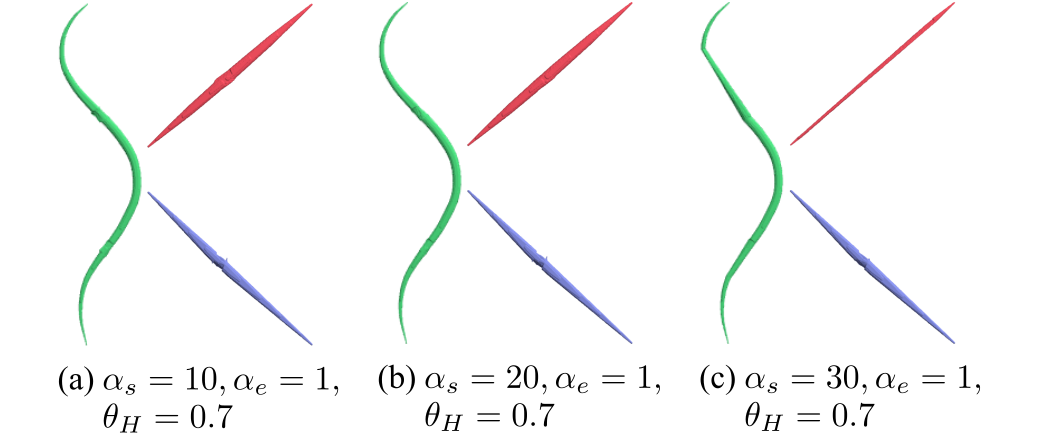}
{Decomposition of the synthetic tubular object at $\alpha_s = 10, 20, 30$ for fixed values of $\alpha_e = 1$ and $\theta_H = 0.7$. We use a linear interpolation to define $\zeta$ and set $\theta_c = 0 ^\circ$. At $\alpha_s = 10$, the decomposition/reconstruction is in agreement with the original object because critical points are detected close to the junctions. Increasing the value of $\alpha_s$ enlarges the decomposition intervals, which may result in inaccuracy while decomposition/reconstruction, e.g., at $\alpha_s = 30$. \label{fig:alpha_s}}

\Figure[t][width=0.99\columnwidth]{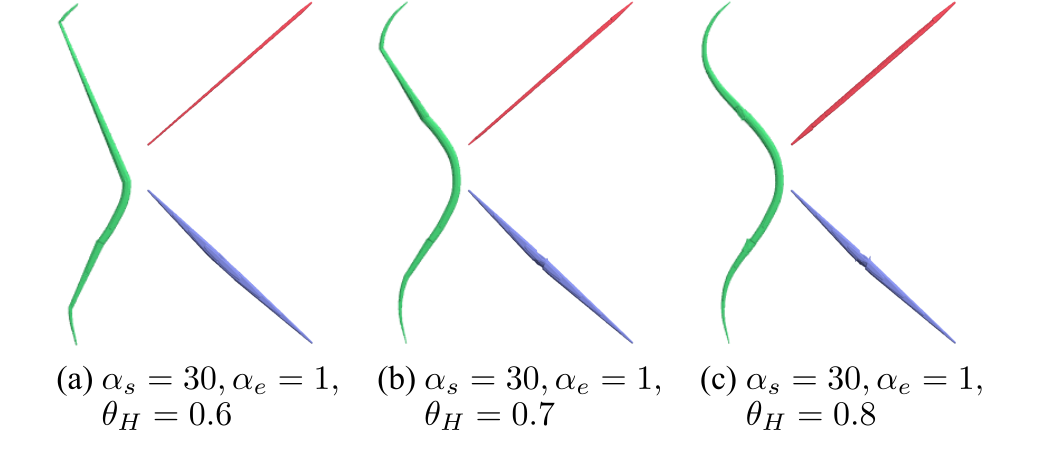}
{Decomposition of the synthetic tubular object at $\theta_H = 0.6, 0.7, 0.8$ for fixed values of $\alpha_s = 30$ and $\alpha_e = 1$. We use a linear interpolation to define $\zeta$ and set $\theta_c = 0 ^\circ$. Increasing the $\theta_H$ value increases the CSD tolerance in dealing with gradient cross-sectional changes of the tubes. At $\theta_H = 0.8$, CSD recognizes the critical points near to junction-points, despite distant starting points from the junctions. \label{fig:theta_H}}

In the skeleton partitioning section (section \ref{sec:skel_partition}), we showed that we merge two successive edges when the angle between them is bigger than $\theta_c$. Therefore, by setting $\theta_c$ to big values, we emphasize the straightness of a path, but then the path may not be maximal-length. Fig. \ref{fig:constraint_ang} shows how $\theta_c$ affects the number of semantic components. At $\theta_c  = 0 ^\circ$, all successive edges are allowed to merge even with acute angles; therefore we obtain a minimum number of object partitions with maximal-length paths (Fig. \ref{fig:constraint_ang}b). By increasing $\theta_c$, only successive edges with a close-to-straight angle are allowed to be merged, which reduces the number of merges and increases the number of object partitions. Fig. \ref{fig:constraint_ang}c shows that decomposition for $\theta_c = 135 ^\circ$ yields four semantic components. At $\theta_c = 180 ^\circ$, no successive edges are merged, and every edge in the skeleton graph corresponds with an object part. Setting $\theta_c > 180 ^ \circ$ generates the maximum number of object parts, equal to the number of skeleton branches (Fig. \ref{fig:constraint_ang}d).

\Figure[t][width=0.99\columnwidth]{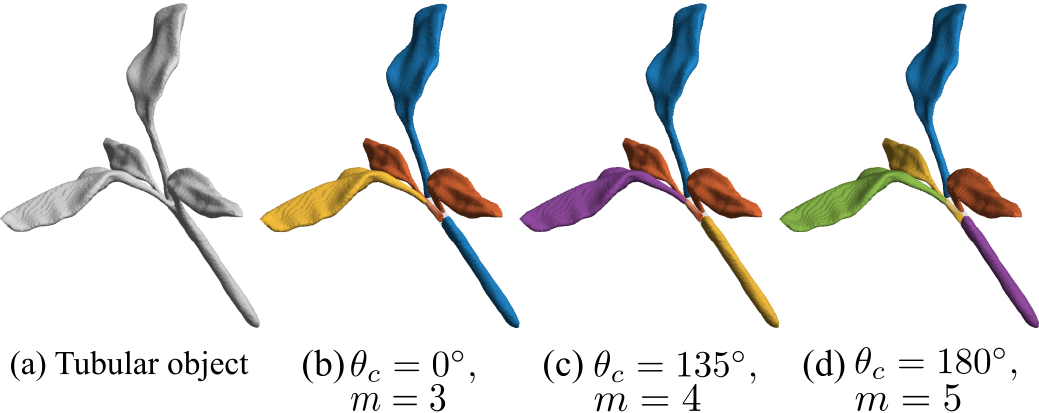}
{The angle between two successive edges in a path should be bigger than $\theta_c$ to be merged. (\textbf{a}) A synthetic tubular object, size: $128 \times 128 \times 128$ voxels. (\textbf{b}) Setting $\theta_c  = 0 ^\circ$ produces maximal-length paths; the minimum number of object parts $m = 3$. (\textbf{c}) At $\theta_c = 135 ^\circ$, the number of object parts increases to $m = 4$. (\textbf{d}) At $\theta_c = 180 ^\circ$, every edge in the skeleton graph is a path, hence producing the maximum number of semantic components, which is equal to the number of skeleton branches, $m = 5$. \label{fig:constraint_ang}}

\Figure[t][width=0.99\columnwidth]{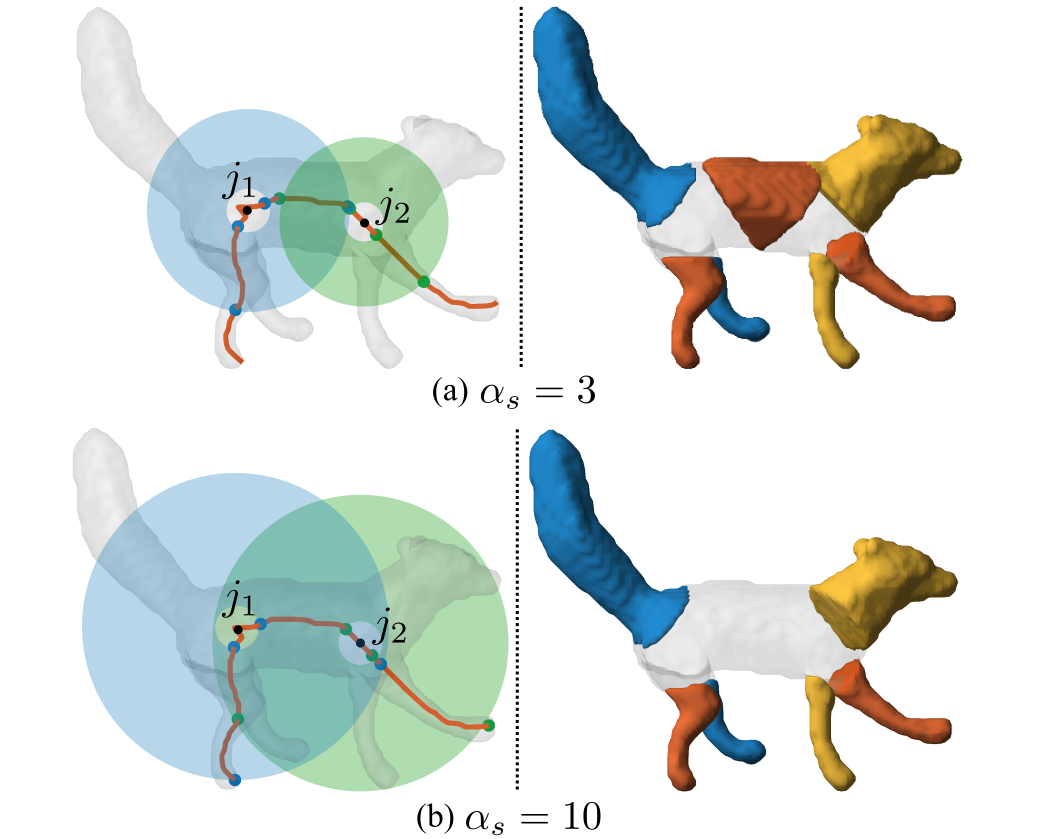}
{Overlapping of decomposition intervals when junction-points are adjacent. (\textbf{a}) Two junction-points (left panel) and two intersections of parts (right panel, grey sections) when $\alpha_s$ is equal to 3. (\textbf{b}) Two junction-points (left panel) and one intersection of parts (right panel, grey section) when $\alpha_s$ is equal to 10. \label{fig:ov_junc}}

We design CSD to have the same number of part intersections as the number of junction-points. However, when two junction-points appear adjacent on a sub-skeleton, we can merge them into one, depending on the value of $\alpha_s$. Fig. \ref{fig:ov_junc}a shows a four-leg with two intersections. For $\alpha_s = 3$, the decomposition interval around $j_1$ does not cover $j_2$, and the decomposition interval around $j_2$ does not cover $j_1$; therefore, the back of the four-leg, within its body, is decomposed. Fig. \ref{fig:ov_junc}b shows that for $\alpha_s = 10$, the decomposition interval around $j_1$ includes $j_2$, and the decomposition interval around $j_2$ includes $j_1$, resulting in one object part intersection, where the four-leg body is the intersection of object parts.

\subsection{Axon segmentation in electron microscopy volumes}

\Figure[tp][width=0.95\textwidth]{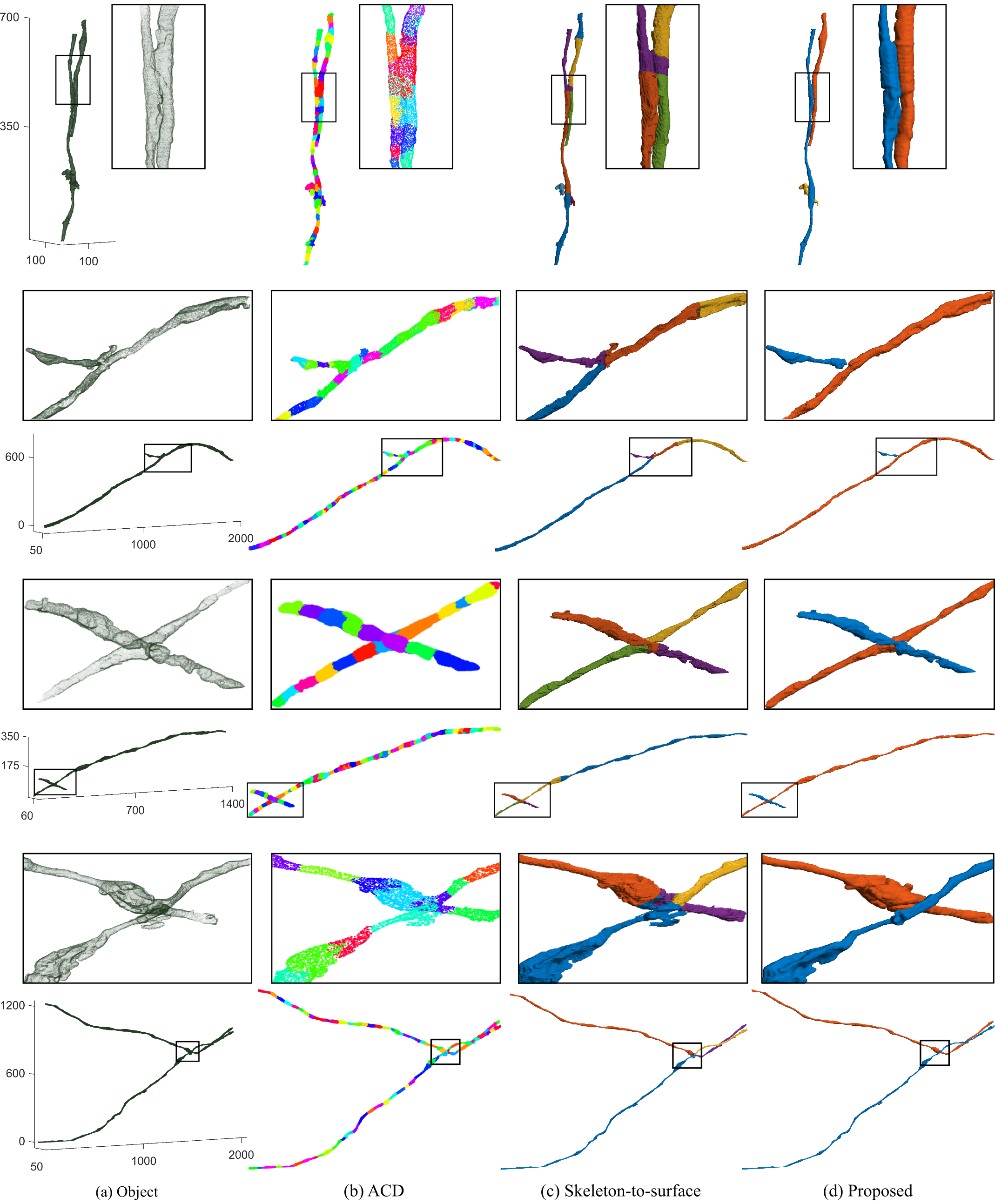}
{(\textbf{a}) Examples of foreground segmentation of myelinated axons with under-segmentation. (\textbf{b}) Decomposition using ACD \cite {Kaick2015ShapeAnalysis}. The point cloud representation of objects is first down-sampled to $50\,000$ points to enable the decomposition task in a reasonable time; this method over-segments the objects. (\textbf{c}) Skeleton-to-surface mapping \cite{Reniers2008ComputingMeasure} based on Voronoi partitioning of the surface using skeleton branches. (\textbf{d}) CSD decomposition provides the correct number of semantic components in under-segmented myelinated axons. The objects are reconstructed at intersections using generalized cylinders. Objects inside boxes are magnified. \label{fig:dec_axons}}

\Figure[t][width=1\textwidth]{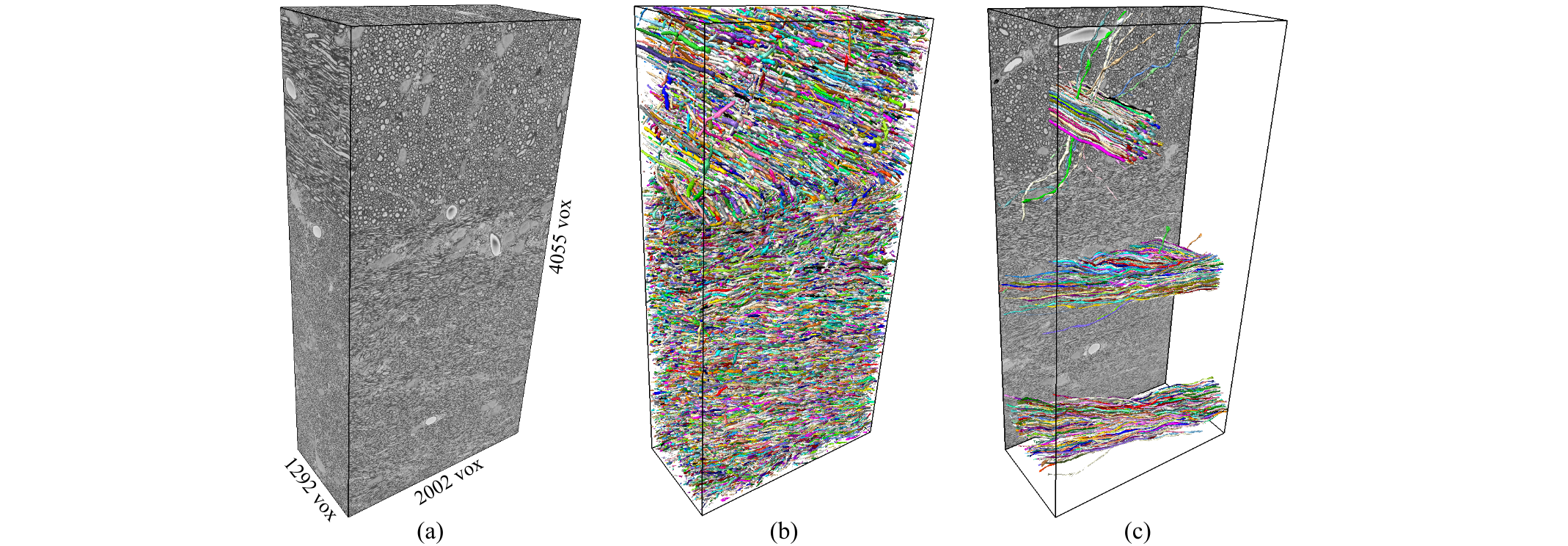}
{(\textbf{a}) A large electron microscopy volume of the white matter. The size of the volume is $4\,055 \times 2\,002 \times 1\,292$ voxels in $x$, $y$, and $z$ directions, respectively. (\textbf{b}) A 3D rendering of myelinated axons (at one-third of the original resolution). CSD evaluates a preliminary segment for under-segmentation error(s), and if required, decomposes and reconstructs an under-segmented myelinated axon. (\textbf{c}) A 3D rendering of myelinated axons sampled at different locations illustrating the diversity of thickness and orientation in segmented axons. \label{fig:whole_vol}}

\Figure[t][width=1\textwidth]{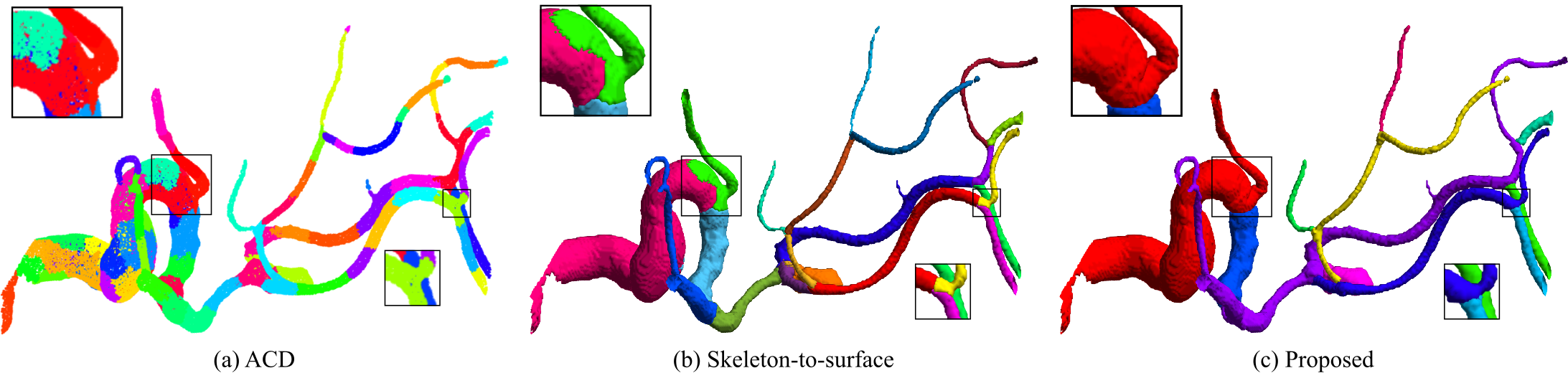}
{(\textbf{a}) ACD \cite {Kaick2015ShapeAnalysis} over-segments the vascular network. (\textbf{b}) Skeleton-to-surface mapping \cite{Reniers2008ComputingMeasure} decomposes the object into 20 semantic components, and the boundaries between these components are not accurate. (\textbf{c}) CSD decomposes the object into eight semantic components and reconstructs the object at intersections. Objects inside boxes are magnified. The 3D image of the vascular network is acquired from Colin Macdonald's GitHub page. \label{fig:vessel}}

\Figure[t][width=0.97\textwidth]{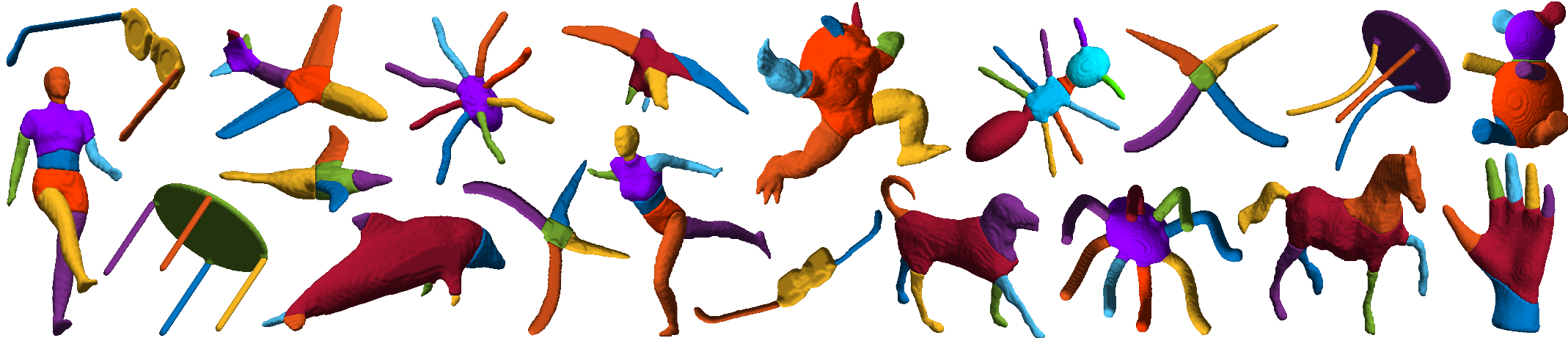}
{A gallery of CSD decomposition of synthetic objects. \label{fig:general_objs}}

The primary purpose of developing CSD is to segment tens of thousands of myelinated axons in electron microscopy volumes of white matter, whose sizes are approximately $4000 \times 2000 \times 1300$ voxels. We generate a probability map of myelinated axons using deep convolutional neural networks (for details, we refer to \cite{Abdollahzadeh2021DeepACSONMicroscopy}). We threshold the probability map, and using connected component analysis, we obtain a preliminary foreground segmentation of myelinated axons. Fig. \ref{fig:dec_axons}a shows examples of myelinated axons after connected component analysis with an under-segmentation error(s): an axon intersects other axons or merges with the extra-axonal space. We apply CSD to evaluate every preliminary segmentation of myelinated axons for the under-segmentation error. If CSD recognizes an under-segmentation error, it decomposes the segmented component into its semantic parts. Fig. \ref{fig:dec_axons} shows the proposed decomposition of myelinated axons compared to the ACD (developed for point clouds) and skeleton-to-surface mapping approaches. To apply ACD on large objects, we first down-sample the point cloud representation of objects to $50\,000$ points, enabling the decomposition to be performed in a reasonable time (less than 10 minutes per object). Fig. \ref{fig:dec_axons}b shows that ACD over-segments myelinated axons. We perform skeleton-to-surface mapping decomposition based on the Voronoi partitioning of surfaces, using Euclidean distance to skeleton branches (Fig. \ref{fig:dec_axons}c). Because a curve skeleton captures the object geometry, skeleton-to-surface mapping decomposes an object close-to-semantic, but it does not recognize intersections of object parts and the boundary cuts are not correct. Fig. \ref{fig:dec_axons}d shows our decomposition of myelinated axons, where CSD generates the correct number of semantic parts for under-segmented myelinated axons and reconstructs axons at intersections using generalized cylinders. Fig. \ref{fig:whole_vol} shows the complete segmentation of myelinated axons in a large electron microscopy volume, where CSD scans, decomposes, and reconstructs about $30\,000$ myelinated axons.

\subsection{Decomposition of vascular networks}
We compare our method to ACD and skeleton-to-surface mapping for the decomposition of a vascular network. Fig. \ref{fig:vessel}a shows that ACD over-segments the vascular network. Fig. \ref{fig:vessel}b shows that skeleton-to-surface mapping decomposes the object into 20 semantic components based on the Euclidean distance to skeleton benches, but the method does not identify intersections, yet the boundary cuts are not correct. For example, Fig. \ref{fig:vessel}b (magnified box) shows that where the thin vessel (green partition) bends on the thick vessel (red partition), skeleton-to-surface mapping erroneously assigns a part of the thick vessel to the thin vessel, the part which is closer to the skeleton of the thin vessel. CSD decomposes the object into eight semantic components and reconstructs the object at intersections (Fig. \ref{fig:vessel}c).

\subsection{Decomposition of synthetic objects}
To demonstrate the general applicability of the CSD algorithm, we examine the proposed CSD method on synthetic voxelized objects. The synthetic objects are from the Princeton segmentation benchmark database \cite{Chen:2009:ABF}. We voxelize meshes from the Princeton database using a ray intersection method described in \cite{Patil2005Voxel-basedShapes}. The resolution of a voxelized object is determined using the bounding box of its OFF model; the bounding box values are normalized to range in $(0, 1]$ then multiplied by 128. The resolution at each dimension is proportional to the length of the bounding box at that dimension, e.g., the dimension with the maximum length is represented by 128 voxels. Fig. \ref{fig:general_objs} shows a gallery of decomposition on a mixture of objects with articulating parts, such as humans, octopuses, or pliers, and objects with moderate or small articulation, such as birds or fishes. Objects such as tables or airplanes include flat parts, which cannot be considered tubular. In Fig. \ref{fig:qual_comp}, we qualitatively compare several of the CSD results to how humans decompose an object into functional parts (the darker the seam, the more people have chosen a cut along that edge \cite{Chen:2009:ABF}). For quantitative analysis, we compare ACD, GCD, and the proposed CSD method to the human decomposition over objects acquired from the Princeton database. We select two objects per category from the Princeton database, excluding categories that have objects with the genus bigger than zero, such as cups or vases, and categories that have an ambiguous skeleton, such as busts or mechs. The objects and their corresponding human segmentations are converted from mesh to voxel-based representation. To aggregate evaluation metrics over multiple human segmentations for the same model and multiple models for the same object category, we report averages per category (averages are computed first within each model, then the results are averaged within each object category). We report Rand error \cite{Rand1971ObjectiveMethods} and boundary error \cite{Martin2001AStatistics} and propose using the variance of information (VOI), which has a better discriminative error range than Rand error \cite{Nunez-Iglesias2013MachineImages}. Table \ref{tbl:quant_syn} shows that the proposed CSD algorithm outperforms both ACD and GCD methods. We stress that these results refer to the decomposition of the voxel-based objects (as obtained in biomedical imaging experiments), and we make no claims about the superiority of CSD when, for example, mesh-based representation of the surfaces would be the natural one. 

\Figure[t][width=0.99\columnwidth]{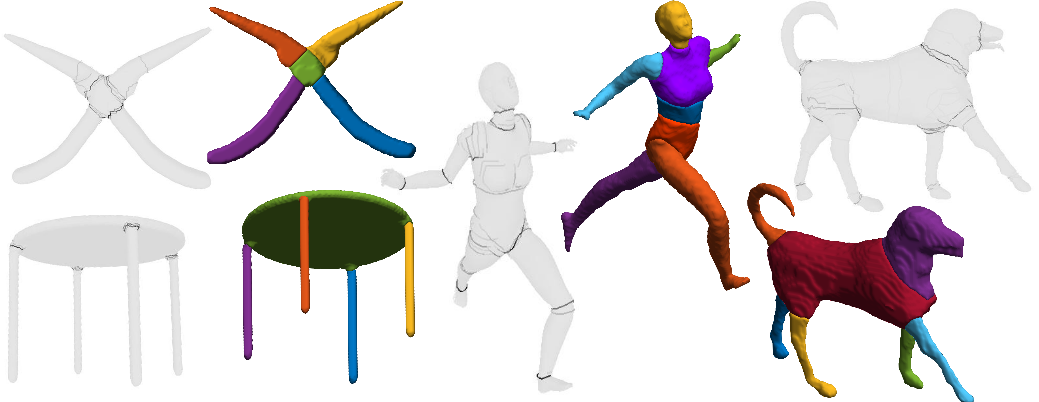}
{Human shape decomposition compared to the CSD decomposition results. The human decomposition of meshes into functional parts are treated as probabilistic ground truth; darker lines show places where more people placed a segmentation boundary. \label{fig:qual_comp}}

\begin{table}[p]
\centering
\caption{Comparison of decomposition techniques using Rand error (RE), the variance of information (VOI), and boundary error (BE) to human shape decomposition; smaller values are better. The average decomposition time (Time) of all objects reported in this table is presented as mean $\pm$ standard deviation. These decomposition techniques are implemented using different programming languages (ACD \cite {Kaick2015ShapeAnalysis}: C++ and Matlab, GCD \cite{Zhou2015GeneralizedDecomposition}: C++, and CSD: Python) and take different object representations as input (ACD: point cloud, GCD: mesh, and CSD: voxel).}

\setlength{\tabcolsep}{3pt}
\renewcommand{\arraystretch}{1.15} 
\begin{tabular}{p{40pt}p{40pt}p{40pt}p{40pt}p{40pt}}
\hline

Input model & Evaluation & ACD \cite{Kaick2015ShapeAnalysis} & GCD \cite{Zhou2015GeneralizedDecomposition} & Proposed \\  \hline
    
            & RE    & 0.2421 & 0.2897 & \textbf{0.2029} \\
Human       & VOI   & 3.0345 & \textbf{1.7049} & 1.7360 \\
            & BE    & 0.4002 & 0.4145 & \textbf{0.1703} \\ \hline
       
            & RE    & 0.2402 & \textbf{0.1650} & 0.2394 \\
Glasses     & VOI   & 1.7264 & \textbf{0.7111} & 0.8842 \\
            & BE    & 0.0654 & 0.0506 & \textbf{0.0088} \\ \hline

            & RE    & 0.2588 & \textbf{0.1450} & 0.1868 \\
Airplane    & VOI   & 1.9981 & \textbf{0.9944} & 1.4765 \\
            & BE    & 0.1066 & 0.0678 & \textbf{0.0677} \\ \hline
 
            & RE    & 0.2284 & \textbf{0.0699} & 0.0842 \\
Ant         & VOI   & 2.4227 & 1.0315 & \textbf{0.5978} \\
            & BE    & 0.4930 & 0.1573 & \textbf{0.0281} \\ \hline            

            & RE    & 0.3135 & 0.1657 & \textbf{0.0215} \\
Octopus     & VOI   & 3.0222 & 0.9684 & \textbf{0.2418} \\
            & BE    & 0.5433 & 0.1736 & \textbf{0.0223} \\ \hline

            & RE    & 0.0353 & 0.0444 & \textbf{0.0109} \\
Table       & VOI   & 0.4656 & 0.2476 & \textbf{0.1082} \\
            & BE    & 0.1530 & 0.0305 & \textbf{0.0051} \\ \hline

            & RE    & 0.3721 & 0.3530 & \textbf{0.0621} \\
Teddy       & VOI   & 3.2006 & 1.1708 & \textbf{0.5542} \\
            & BE    & 1.5860 & 0.2210 & \textbf{0.1851} \\ \hline 
 
            & RE    & 0.3625 & 0.3400 & \textbf{0.2201} \\
Hand        & VOI   & 2.7837 & \textbf{1.6841} & 1.1728 \\
            & BE    & 0.5442 & 0.2967 & \textbf{0.1739} \\ \hline 

            & RE    & 0.2399 & 0.0830 & \textbf{0.0656} \\
Pliers      & VOI   & 2.7210 & 0.6070 & \textbf{0.5419} \\
            & BE    & 0.2736 & 0.0936 & \textbf{0.0386} \\ \hline

            & RE    & 0.5879 & 0.5140 & \textbf{0.2232} \\
Fish        & VOI   & 2.6352 & 1.7495 & \textbf{0.8839} \\
            & BE    & 1.1104 & 0.4107 & \textbf{0.1337} \\ \hline

            & RE    & 0.1764 & 0.2309 & \textbf{0.2208} \\
Bird        & VOI   & 1.6971 & \textbf{1.2629} & 1.3959 \\
            & BE    & 1.0669 & 0.1431 & \textbf{0.0901} \\ \hline

            & RE    & 0.2312 & 0.1897 & \textbf{0.1870} \\
Armadillo   & VOI   & 2.9909 & \textbf{1.3408} & 1.5887 \\
            & BE    & 0.8717 & 0.3583 & \textbf{0.3407} \\ \hline
    
            & RE    & 0.4082 & 0.3667 & \textbf{0.1868} \\
Four-leg     & VOI   & 2.9593 & 1.9207 & \textbf{1.0821} \\
            & BE    & 0.6796 & 0.3486 & \textbf{0.1738} \\ \hline \hline
            
            & Time & 1287 $\pm$ 820 & 336 $\pm$ 98 & \textbf{186 $\pm$ 42} \\ \hline
 
\end{tabular}
\label{tbl:quant_syn}
\end{table}

\subsection{Decomposition of noisy synthetic objects}
We develop CSD to decompose voxel-based objects, e.g., objects extracted from biomedical images, where noise can degrade the object surface. To examine how noise affects decomposition techniques, as shown in Fig. \ref{fig:noise_dec_tech}a, we add impulse noise to the surface of an object for different noise density $D_n$ values; $D_n = 0, 0.1, 0.35$, and 0.6. Fig. \ref{fig:noise_dec_tech}b shows that ACD over decomposes the noise-free object to 64 parts and 82 parts when noise density equals 0.6. The GC decomposition of the noise-free object is approximately correct, but over-decomposes the noisy object at $D_n = 0.1$ (Fig. \ref{fig:noise_dec_tech}c). GCD does not decompose the object at stronger noise levels when $D_n$ equals 0.3 or 0.6. The proposed CSD method decomposes the objects at different noise levels with excellent performance, as shown in Fig. \ref{fig:noise_dec_tech}f. For the quantitative analysis, we compare decompositions against human segmentation using Rand error, VOI, and boundary error, as in Table \ref{tbl:noise_dec_tech}. The ACD and GCD decomposition errors are high on different metrics while constantly low for the proposed CSD method.

\Figure[t][width=0.99\textwidth]{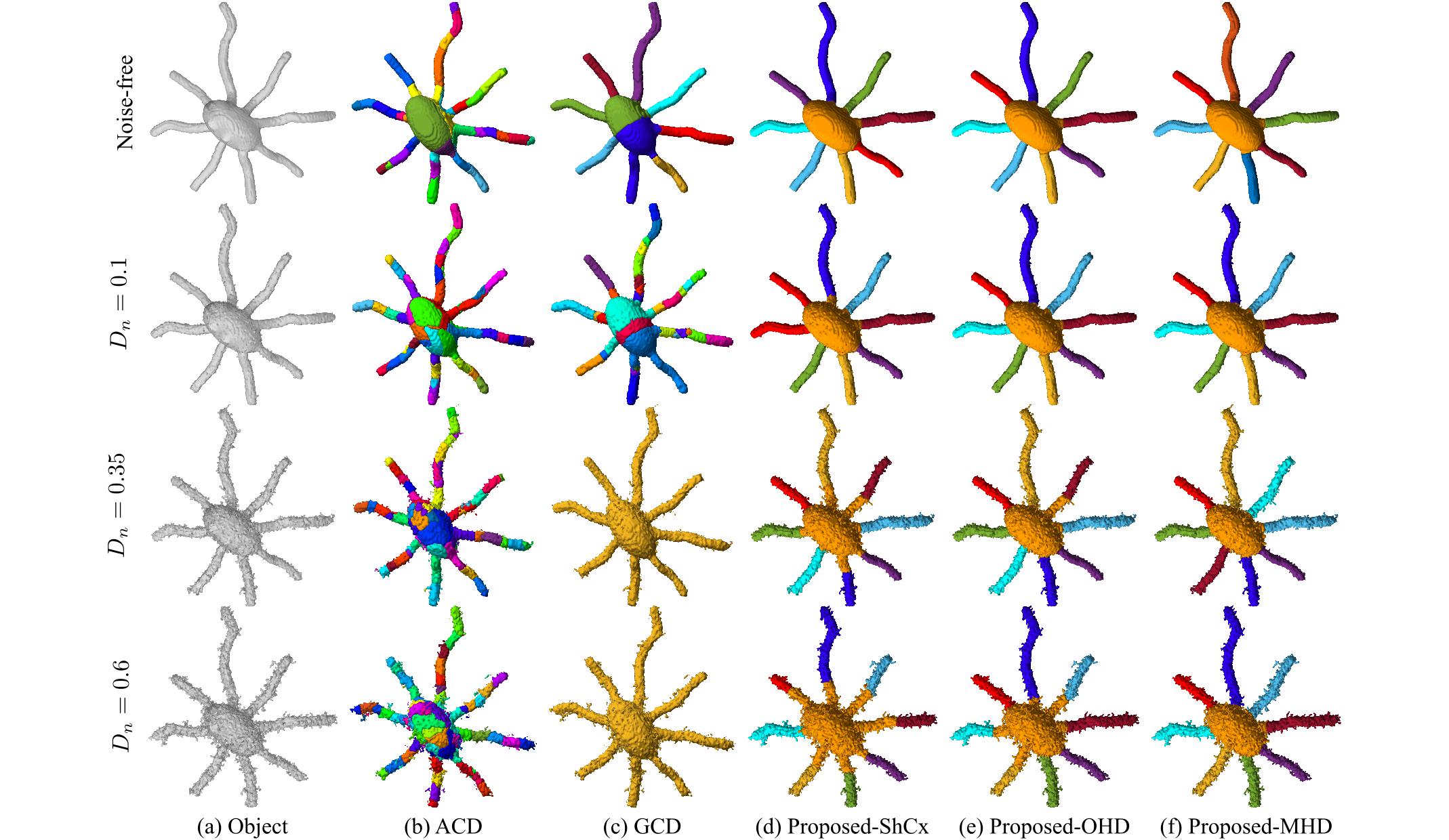}
{Decomposition of (\textbf{a}) ACD \cite{Kaick2015ShapeAnalysis}, (\textbf{b}) GCD \cite{Zhou2015GeneralizedDecomposition}, and the proposed CSD method (\textbf{d-f}) when the object surface is degraded with the impulse noise for the noise density $D_n$ equals 0, 0.1, 0.35, and 0.6. The original Hausdorff distance as the similarity measure in the proposed CSD method is substituted with alternative metrics: (\textbf{d}) the shape context (ShCx) method \cite{Belongie2002ShapeContexts}, (\textbf{e}) original Hausdorff distance (OHD), and (\textbf{f}) modified Hausdorff distance (MHD) \cite{Dubuisson2002AMatching}. \label{fig:noise_dec_tech}}

\begin{table*}[t]
\centering
\caption{Comparison of different techniques over the voxel-based object shown in Fig. \ref{fig:noise_dec_tech} using Rand error (RE), the variance of information (VOI), and boundary error (BE). Smaller values indicate a closer decomposition to how humans decompose an object into functional parts. The average decomposition time (Time) of all objects reported in this table is presented as mean $\pm$ standard deviation. The GCD \cite{Zhou2015GeneralizedDecomposition} method does not decompose objects at strong noise levels when $D_n$ equals 0.3 or 0.6 and returns the object itself; therefore, we reported the GCD decomposition time for all experiments separately, showing the decomposition time of a failure case with $\infty$. These decomposition techniques are implemented using different programming languages (ACD \cite {Kaick2015ShapeAnalysis}: C++ and Matlab, GCD: C++, and CSD: Python) and take different object representations as input (ACD: point cloud, GCD: mesh, and CSD: voxel).}

\setlength{\tabcolsep}{3pt}
\renewcommand{\arraystretch}{1.15} 
\begin{tabular}{p{60pt}p{60pt}p{60pt}p{60pt}p{60pt}p{60pt}p{60pt}}
\hline
    Noise level  & Evaluation & ACD \cite{Kaick2015ShapeAnalysis} & GCD \cite{Zhou2015GeneralizedDecomposition}  & Proposed-ShCx & Proposed-OHD & Proposed-MHD \\  \hline
    
                 & RE    & 0.3633 &   0.2659 & 0.0276 & \textbf{0.0211} & 0.0263 \\
    Noise free   & VOI   & 2.9877 &   1.2150 & 0.3046 & \textbf{0.2482} & 0.2940 \\
                 & BE    & 0.5377 &   0.1976 & 0.0213 & \textbf{0.0157} & 0.0204 \\ \hline
    
                 & RE    & 0.4159 &   0.3021 & 0.0530 & \textbf{0.0207} & 0.0249 \\
    $D_n$ = 0.1  & VOI   & 3.5099 &   1.9760 & 0.5486 & \textbf{0.2424} & 0.2819 \\
                 & BE    & 0.7263 &   0.4015 & 0.0795 & \textbf{0.0681} & 0.0705 \\ \hline
    
                 & RE    & 0.4677 &   0.4951 & 0.1064 & 0.0432 & \textbf{0.0360} \\
    $D_n$ = 0.35 & VOI   & 3.7282 &   1.7636 & 0.7965 & 0.3702 & \textbf{0.3688} \\
                 & BE    & 1.1797 &   0.7403 & 0.2295 & \textbf{0.2248} & 0.2304 \\ \hline
                 
                 & RE    & 0.4221 &   0.5567 & 0.2036 & 0.0999 & \textbf{0.0250} \\
    $D_n$ = 0.6  & VOI   & 3.9624 &   1.9533 & 1.2451 & 0.7191 & \textbf{0.2787} \\
                 & BE    & 0.9429 &   0.4605 & 0.3615 & 0.3581 & \textbf{0.3569} \\ \hline \hline
                 
                 & Time & 1302 $\pm$ 438 & [360, 482, $\infty$, $\infty$] & 1021 $\pm$ 468  & \textbf{191 $\pm$ 26} & 193 $\pm$ 25 \\ \hline
                
\end{tabular}
\label{tbl:noise_dec_tech}
\end{table*}

\subsection{Cross-sectional similarity metric}
To define a critical point in section \ref{sec:critical_point}, we use the Hausdorff distance as defined in \eqref{eq:HausD} to compare geometrical changes between cross-sectional contours. The Hausdorff distance can be sensitive to surface noise, showing a mismatch between cross-sectional contours that belong to the same object part. Figs. \ref{fig:noise_dec_tech}d-f show the CSD decomposition performance, substituting the original Hausdorff distance with alternative shape matching techniques: the modified Hausdorff distance \cite{Dubuisson2002AMatching} and shape context metric \cite{Belongie2002ShapeContexts}. We set $\theta_H$ equal to 0.8 for the original and modified Hausdorff distances and set a similarity threshold of 0.5 for the shape context metric. The surface of the synthetic object in Fig. \ref{fig:noise_dec_tech}a is degraded with the impulse noise for the noise density $D_n$ equals 0, 0.1, 0.35, and 0.6. In Table \ref{tbl:noise_dec_tech}, we quantitatively evaluate decompositions using different cross-sectional similarity metrics and at different noise levels to the human segmentation using Rand error, VOI, and boundary error. Results demonstrate that different CSD similarity metrics yield excellent decompositions for the noise-free object, where the original Hausdorff distance performs better than the shape context metric and modified Hausdorff distance. With increasing the noise density, however, modified Hausdorff distance performs better than the shape context and the original Hausdorff metrics. The modified Hausdorff distance produces an excellent decomposition across noise densities.

\subsection{Choice of skeletonization technique}
The skeleton partitioning step guides the CSD algorithm for semantic decomposition; therefore, the quality of the skeletonization itself is crucial for the decomposition. We compare the CSD skeletonization approach to the L1-medial skeletonization and ROSA techniques in terms of the topological correctness and centeredness of the extracted skeletons. We compare these skeletonization techniques on synthetic voxel-based objects from McGill 3D Shape Benchmark \cite{Siddiqi2008RetrievingSurfaces}. Fig. \ref{fig:skl_comp} shows that when two object parts appear very close, L1-medial (Fig. \ref{fig:skl_comp}a human hand) and ROSA (Fig. \ref{fig:skl_comp}b octopus) merge the object parts and form a cycle in genus zero objects. The CSD skeletonization, see Fig. \ref{fig:skl_comp}c, generates separate branches for different object parts. Also, flat object parts, such as tabletop, do not possess a reasonably meaningful curve skeleton, but CSD skeletonization generates a more meaningful skeleton than the other two methods. In terms of centeredness, L1-medial and ROSA do not stay within the object. The CSD distance-based skeletonization approach penalizes the skeleton where the path moves away from the center of the object. Table \ref{tbl:skl_time} shows the computation time spent on skeletonization with L1-medial, ROSA, and CSD methods. The CSD skeletonization is faster than the other two methods, rendering it suitable for large objects.

\Figure[t][width=1\textwidth]{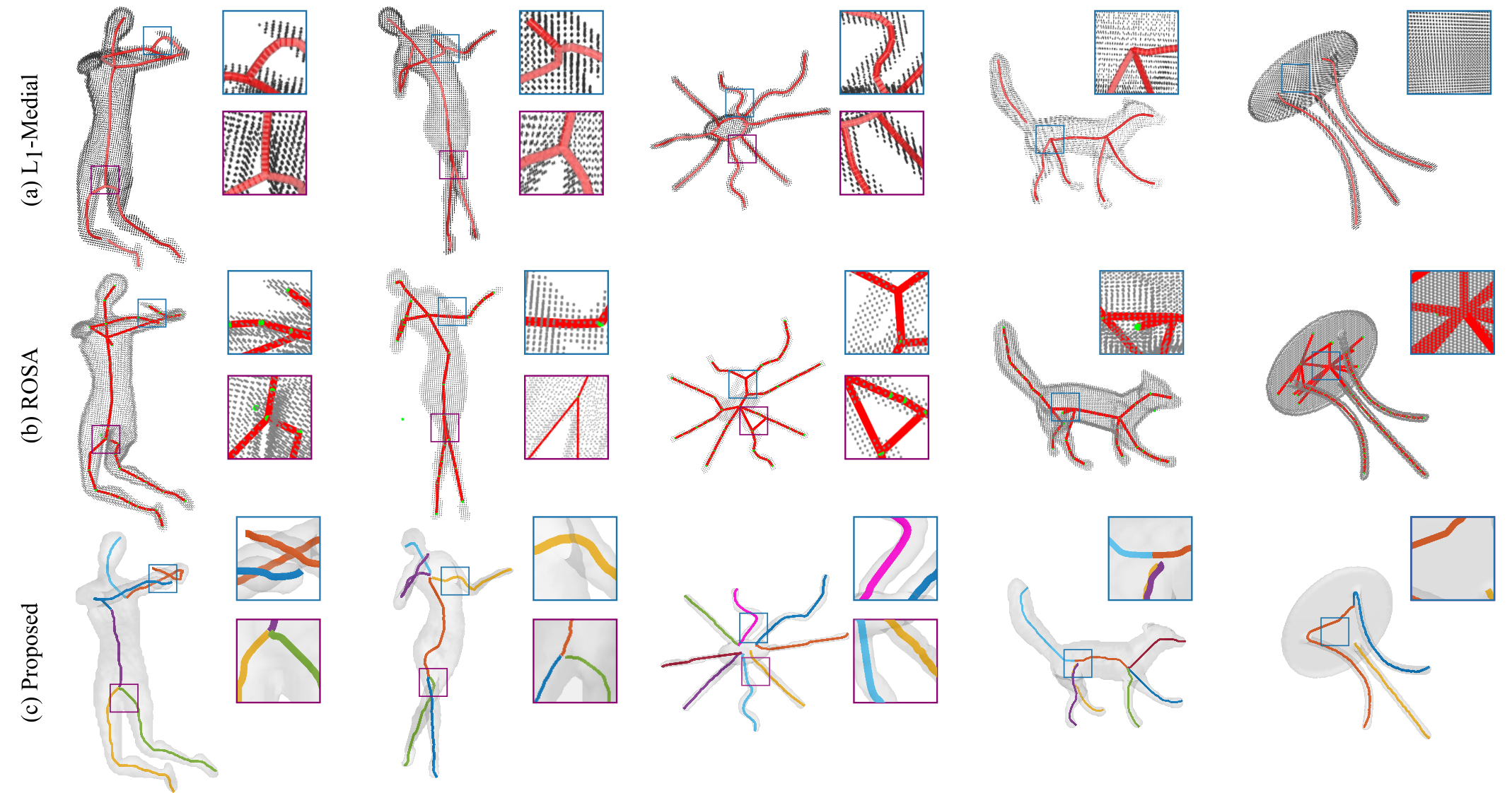}
{Skeletonization using (\textbf{a}) L1-medial \cite{Huang2013LCloud}, (\textbf{b}) ROSA \cite{Tagliasacchi2009CurveCloud}, and (\textbf{c}) CSD distance-based techniques. L1-medial and ROSA may incorrectly form cycles in the skeleton of genus zero objects when two object parts are close, while the CSD skeletonization generates separate branches for different object parts. Also, L1-medial and ROSA do not necessarily stay within the object, while the CSD distance-based skeletonization approach penalizes the skeleton where the path moves away from the center of the object. Objects inside boxes are magnified. \label{fig:skl_comp}}

\begin{table}
\setlength{\tabcolsep}{3pt}
\renewcommand{\arraystretch}{1.2} 
\centering
\caption{The skeletonization time for objects in Fig. \ref{fig:skl_comp} using L1-medial \cite{Huang2013LCloud}, ROSA \cite{Tagliasacchi2009CurveCloud}, and the proposed CSD method. These methods are implemented using different programming languages (L1-medial: C++, ROSA: C++ and Matlab, and CSD: Python). L1-medial and ROSA take point clouds as input, while CSD takes voxel-based objects as input. Therefore, the times reported are not directly comparable but provide insight into the speed of skeletonization algorithms, which is important for our segmentation application of large objects. For L1-medial and ROSA, we report the number of points (pts). For CSD, we report the number of voxels representing objects; the bounding box of objects is $128 \times 128 \times 128$ voxels.}
\begin{tabular}{p{30pt}p{35pt}p{50pt}p{50pt}p{40pt}}
\hline
    Model   & Evaluation & L1-medial \cite{Huang2013LCloud}   & ROSA \cite{Tagliasacchi2009CurveCloud}     & Proposed      \\  \hline
    
    \multirow{2}{*}{Human}  & Size     & $1\,000$ pts & $13\,189$ pts & $30\,785$ vox \\ 
            & Time (s) & 37           & 549           & \textbf{4}    \\  \hline
                   
    \multirow{2}{*}{Human}   & Size     & $1\,000$ pts & $11\,107$ pts & $24\,897$ vox \\
            & Time (s) & 31           & 403           & \textbf{4}    \\ \hline
              
    \multirow{2}{*}{Octopus} & Size     & $1\,000$ pts & $6\,415$  pts & $9\,179$ vox  \\
            & Time (s) & 19           & 144           & \textbf{6}    \\ \hline
              
    \multirow{2}{*}{Four-leg} & Size   & $1\,000$ pts & $11\,878$ pts & $30\,490$ vox \\
              &Time (s)& 31           & 465           & \textbf{4}    \\ \hline 
              
    \multirow{2}{*}{Table}     & Size   & $1\,000$ pts & $19\,499$ pts & $42\,323$ vox \\
              &Time (s)& 80           & $1\,333$      & \textbf{4}    \\ \hline              
\end{tabular}
\label{tbl:skl_time}
\end{table}

\subsection{Computation time}
The time complexity of the sub-voxel precise skeletonization is $O(n \, N_\Omega \log N_\Omega)$, where $n$ is the number of skeleton branches, and $N_\Omega$ is the number of voxels in a discrete $\Omega$. The $N_\Omega \log N_\Omega$ factor is from the fast marching algorithm \cite{Sethian1996AFronts}. The time complexity to determine a critical point is $O(N_p)$, where $N_p$ is the number of inquiry points that we check for the cross-sectional changes in a decomposition interval. Defining the critical points is independent of $N_\Omega$. The complexity of the method is measured through the number of basic arithmetic operations performed; other factors that may also influence the execution time, such as the number of memory accesses or memory consumption, have not been considered.

A fair comparison between the computation times of different decomposition techniques by the wall clock time requires the same constraints for all techniques. Providing such constraints is challenging because different decomposition techniques use different programming languages or take different object representations as input. In spite of that, we demonstrate the average decomposition time of ACD, GCD, and the proposed CSD method in Tables \ref{tbl:quant_syn} and \ref{tbl:noise_dec_tech}. The average decomposition time of the proposed CSD is shorter than ACD and GCD in all our experiments: the average decomposition time of synthetic objects is 3 m for CSD, 21 m for ACD, and 5 m for GCD. Note that, although the computation times of the GCD and proposed CSD methods are close, we have not been able to decompose objects acquired from biomedical image datasets, e.g., axons (Fig. \ref{fig:dec_axons}) and the vascular network (Fig. \ref{fig:vessel}), using GCD within a day. 

We also propose to reduce the CSD computation time by reducing the number of inquiry points $N_p$. For that, we propose to sub-sample the sub-skeletons by a sampling factor $sf$, as shown in Fig. \ref{fig:sf}. Increasing $sf$ reduces the computation time linearly while affecting decomposition results minimally. Fig. \ref{fig:sf} shows the decomposition of three objects at $sf$ equals 1, 4, 16, and 64. To evaluate the effect of sampling the sub-skeletons on decomposition results, we compared decomposition results over Rand error, VOI, and boundary error, considering decomposition at $sf = 1$ as the ground truth. Table \ref{tbl:sf} shows that decomposition by a factor of four substantially reduces the computation time, whereas the evaluation metrics worsen minimally. Reducing the computation is important when dealing with big voxel-based objects. For example, on a 2 $\times$ Intel Xeon E5 $2\,630$ CPU 2.4 GHz machine with 512 GB RAM using Python 3.6, the skeletonization of the myelinated axon shown in the first row of Fig. \ref{fig:dec_axons} ($N_\Omega = 395\,594$) consumes 117 s and defining its critical points 353 s, and sampling the sub-skeletons by $sf = 5$ reduces the decomposition time to 75 s. 

\Figure[t][width=0.99\columnwidth]{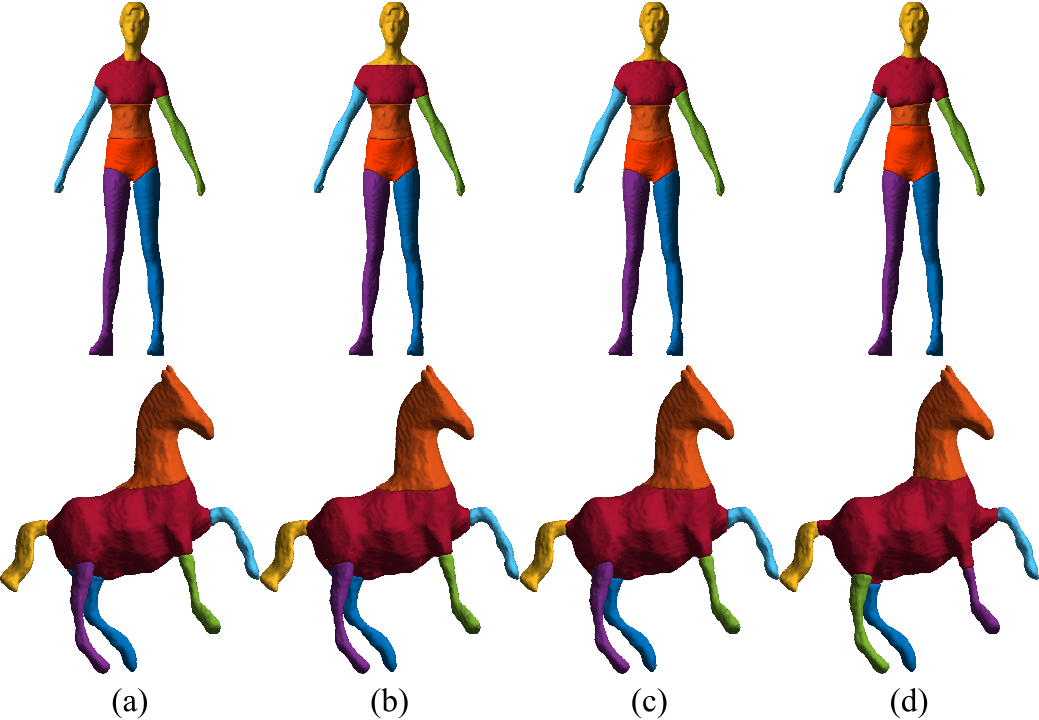}
{Sampling object sub-skeletons to reduce the decomposition time. The decomposition of a human and a four-leg at (\textbf{a}) $sf = 1$, (\textbf{b}) $sf = 4$, (\textbf{c}) $sf = 16$, and (\textbf{d}) $sf = 64$. \label{fig:sf}}

\begin{table}
\caption{Evaluation of how sampling the sub-skeletons for $sf$ equals 1, 4, 16, and 64 affects decomposition results of objects in Fig. \ref{fig:sf} using Rand error (RE), the variance of information (VOI), and boundary error (BE), considering decomposition at $sf = 1$ as the ground truth. The decomposition results are achieved on a 4-core Intel CPU 3.41 GHz computer with 64 GB RAM using Python 3.6.}

\setlength{\tabcolsep}{3pt}
\renewcommand{\arraystretch}{1.2} 
\begin{tabular}{p{35pt}p{35pt}p{33pt}p{33pt}p{33pt}p{33pt}}
\hline
    Model     & Evaluation & $sf = 1$ & $sf = 4$ & $sf = 16$ & $sf = 64$ \\  \hline
    
   \multirow{4}{*}{Human} & RE        &     & \textbf{0.0140} & 0.0203 & 0.0340 \\
              & VOI       &     & \textbf{0.2507} & 0.4082 & 0.5511 \\
              & BE        &     & \textbf{0.0496} & 0.0915 & 0.1797 \\
              & Time (s)  & 108.1 & 30.5     & 9.2      & \textbf{4.5}      \\ \hline
              
    \multirow{4}{*}{Four-leg} & RE         &     & \textbf{0.0319} & 0.0359 & 0.0519 \\
              & VOI        &     & \textbf{0.2149} & 0.2775 & 0.3971 \\
              & BE         &     & \textbf{0.0600} & 0.0652 & 0.0714 \\
              & Time (s)   & 174.0 & 48.3     & 14.1     & \textbf{4.9}      \\\hline 
              
\end{tabular}
\label{tbl:sf}
\end{table}

\section{Conclusion}
In this paper, we proposed the application of 3D shape decomposition in image segmentation. We presented the novel CSD algorithm to decompose and reconstruct under-segmented tubular objects. The CSD method is guided by the curve skeleton decomposition, decomposing a tubular object into maximal-length, approximately straight parts. The object is cut at the intersection of parts using translational sweeps and reconstructed by generalized cylinders. 
We demonstrated the application of CSD on biomedical imaging volumes and synthetic objects. In particular, we applied CSD as instance segmentation to deep learning-based semantic segmentation of myelinated axons. Hundreds of thousands of myelinated axons were automatically evaluated for under-segmentation error, and under-segmented myelinated axons were decomposed into their constituent axons, using the same parameter values for all objects in all electron microscopy datasets. We showed that CSD outperforms state-of-the-art techniques in decomposing voxel-based objects and is robust to severe surface noise. CSD is highly parallelizable, substantially reducing the computation time of the segmentation in large biomedical imaging datasets. The proposed CSD algorithm allows for including the cylindricity as a global shape-objective for a fast 3D segmentation of tubular objects in large biomedical imaging datasets.

\section*{Acknowledgmen}
The authors acknowledge CSC-IT Center for Science, Finland and Bioinformatics Center, University of Eastern Finland, Finland, for computational resources.

\section*{Competing interests}
\noindent The authors declare that they have no conflict of interest.

\section*{Availability of data and material}
\noindent The source code of the CSD algorithm is available at https://github.com/aAbdz/CylShapeDecomposition.

\bibliographystyle{IEEEtran}
\bibliography{CylShapeDec.bib} 

\begin{IEEEbiography}[{\includegraphics[width=1in, height=1.25in, clip, keepaspectratio]{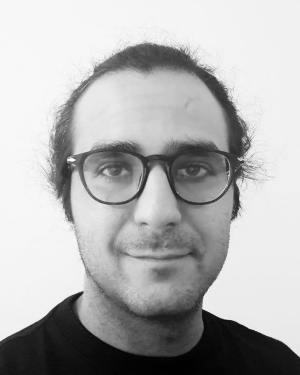}}] {Ali Abdollahzadeh} received his master's degree in Electrical Engineering from Tampere University of Technology, Finland, in 2016. Currently, he is a Ph.D. student at the A.I. Virtanen Institute for Molecular Sciences at the University of Eastern Finland, Kuopio, Finland. His research interests include shape analysis, image processing, machine learning, and their applications in large electron- and light-microscopy image volumes of the brain.
\end{IEEEbiography}

\begin{IEEEbiography}[{\includegraphics[width=1in, height=1.25in, clip, keepaspectratio]{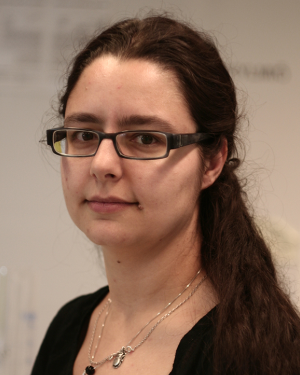}}] {Alejandra Sierra} obtained her Ph.D. degree in Biochemistry from the Autonomous University of Madrid (Spain) in 2006. Since then, she has worked at the A.I. Virtanen Institute for Molecular Sciences at the University of Eastern Finland, Kuopio, Finland. During this time, she held both Academy of Finland post-doctoral and research fellow positions, and now she is a research director in the same institution. Her research interest is the validation and development of magnetic resonance imaging techniques by incorporating microscopic tissue information in the healthy and diseased brain.
\end{IEEEbiography}

\begin{IEEEbiography}[{\includegraphics[width=1in, height=1.25in, clip, keepaspectratio]{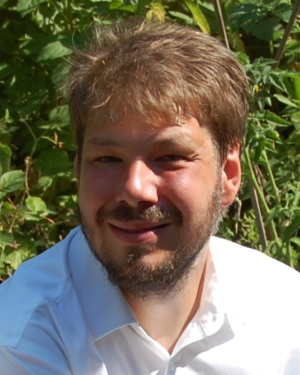}}] {Jussi Tohka} received his Ph.D. degree in Signal Processing from Tampere University of Technology, Finland, in 2003. He was a post-doctoral fellow in the University of California, Los Angeles, USA, held an Academy research fellow position at the Department of Signal Processing, Tampere University of Technology, Finland and was a CONEX professor at the Department of Bioengineering and Aerospace Engineering at Universidad Carlos III de Madrid, Spain. He is currently with the A.I. Virtanen Institute for Molecular Sciences at the University of Eastern Finland, Kuopio, Finland. His research interests include machine learning, image analysis, and pattern recognition and their applications to brain imaging.
\end{IEEEbiography}
\EOD

\end{document}